\documentclass[lettersize,journal]{IEEEtran}

\usepackage{graphicx, subcaption}

\usepackage{amsmath,amsfonts}
\usepackage{algorithmic}
\usepackage{algorithm}
\usepackage{array}
\usepackage[caption=false,font=normalsize,labelfont=sf,textfont=sf]{subfig}
\usepackage{textcomp}
\usepackage{stfloats}
\usepackage{url}
\usepackage{verbatim}
\usepackage{cite}

\usepackage{pifont}

\usepackage{booktabs}
\usepackage{multirow, makecell} 

\usepackage{hhline}

\newcommand{\mycc}{\cellcolor{lavender}}
\usepackage{xcolor,colortbl}
\definecolor{lavender}{rgb}{0.9, 0.9, 0.98}

\begin{document}
	
	\title{A Multi-Prototype-Guided Federated Knowledge Distillation Approach in AI-RAN Enabled Multi-Access Edge Computing System}
	
	\author{Luyao Zou, Hayoung Oh, Chu Myaet Thwal, Apurba Adhikary, Seohyeon Hong, and Zhu Han, ~\IEEEmembership{Fellow,~IEEE}
    \thanks{This work has been submitted to the IEEE for possible publication. Copyright may be transferred without notice, after which this version may no longer be accessible.}
	\thanks{Luyao Zou, Hayoung Oh and Seohyeon Hong are with the College of Computing and Informatics, Sungkyunkwan University, Republic of Korea, (emails: zouluyao@skku.edu, hyoh79@skku.edu, hsally1207@skku.edu).}
	\thanks{Chu Myaet Thwal is with the Department of Computer Science and Engineering, Kyung Hee University, Yongin-si, Gyeonggi-do, 17104, Republic of Korea, (email: chumyaet@khu.ac.kr).}
	\thanks{Apurba Adhikary is with the Department of Information and Communication Engineering, Noakhali Science and Technology University, Noakhali-3814, Bangladesh (e-mail: apurba@nstu.edu.bd).}
	\thanks{Zhu Han is with the Department of Electrical and Computer Engineering, University of Houston, Houston 77004, USA, and also with the Department of Computer Science and Engineering, Kyung Hee University, South Korea, (e-mail: hanzhu22@gmail.com).}
	\thanks{Hayoung Oh is the corresponding author.}
	}



\markboth{Journal of \LaTeX\ Class Files,~Vol.~14, No.~8, August~2021}%
{Shell \MakeLowercase{\textit{et al.}}: A Sample Article Using IEEEtran.cls for IEEE Journals}


\maketitle

\begin{abstract}

With the development of wireless network, Multi-Access Edge Computing (MEC) and Artificial Intelligence (AI)-native Radio Access Network (RAN) have attracted significant attention. Particularly, the integration of AI-RAN and MEC is envisioned to transform network efficiency and responsiveness. Therefore, it is valuable to investigate AI-RAN enabled MEC system. Federated learning (FL) nowadays is emerging as a promising approach for AI-RAN enabled MEC system, in which edge devices are enabled to train a global model cooperatively without revealing their raw data. However, conventional FL encounters the challenge in processing the non-independent and identically distributed (non-IID) data. Single prototype obtained by averaging the embedding vectors per class can be employed in FL to handle the data heterogeneity issue. Nevertheless, this may result in the loss of useful information owing to the average operation. Therefore, in this paper, a \underline{m}ulti-\underline{p}rototype-guided \underline{fed}erated \underline{k}nowledge \underline{d}istillation (MP-FedKD) approach is proposed. Particularly, self-knowledge distillation is integrated into FL to deal with the non-IID issue. To cope with the problem of information loss caused by single prototype-based strategy, multi-prototype strategy is adopted, where we present a conditional hierarchical agglomerative clustering (CHAC) approach and a prototype alignment scheme. Additionally, we design a novel loss function (called LEMGP loss) for each local client, where the relationship between global prototypes and local embedding will be focused. Extensive experiments over multiple datasets with various non-IID settings showcase that the proposed MP-FedKD approach outperforms the considered state-of-the-art baselines regarding accuracy, average accuracy and errors (RMSE and MAE).

\end{abstract}


\section{Introduction}
\IEEEPARstart{M}{ulti-Access} Edge Computing (MEC) and Artificial Intelligence (AI)-native Radio Access Network (RAN) nowadays have attracted widespread attention. Specifically, MEC is capable of bringing computational resources and memory resources to the edge of network, which can achieve low communication latency and provide convenient services to accessed devices \cite{Z_Wu_FedICT_Federated_Multi_Task}. Regarding AI-RAN, as specified by the AI-RAN Alliance, AI-RAN aims to tackle a converged 6G platform capable of orchestrating, managing, and deploying the workloads of both AI and RAN \cite{S_D_A_Shah_Interplay_AI_RAN}. Integrating MEC into the RAN can enable the task-specific AI agents to be executed in close proximity to the data source \cite{Z_Nezami_GenAI_RAN}, which is poised to bring revolution to the network efficiency and responsiveness \cite{K_Srikandabala_Scoping}. In light of this, in this article, AI-RAN enabled MEC system is taken into consideration. However, the data produced by the diverse devices in the MEC scenarios is in a growth trend, as the evolution of the wireless network technology and the widespread adoption of the mobile devices \cite{Z_Wu_FedICT_Federated_Multi_Task}. To cope with the tremendous amount of data, the traditional centralized training manner is impractical because of data privacy concerns caused by the necessity of centralized data gathering at a central server \cite{Y_Cheng_Pre_clustering_CFL_Data_System}. 

\par
In response to this, federated learning (FL) can be considered as a promising technique \cite{K_Wang_Age_of_Information_Minimization_FL} as it follows a distributed learning paradigm. Specifically, with respect to FL, it allows many edge clients to collaboratively train models without local data exposure \cite{Y_Xu_Overcoming_Noisy_Labels_Non_IID_Edge}. Generally, FL contains the following steps: \lowercase\expandafter{\romannumeral1}) the central server (CS) generates an initial global model that will be broadcast to the local clients to initialize each local model; \lowercase\expandafter{\romannumeral2}) then the local models will be trained to produce new local models; \lowercase\expandafter{\romannumeral3}) those newly generated local models will be transmitted back to CS; \lowercase\expandafter{\romannumeral4}) with the received local models, the CS will perform model aggregation to produce new global model. These processes will be iterated until convergence or terminated by the determination of the CS \cite{M_Moshawrab_Securing_FL_Approaches}. 
\par
	Nevertheless, deploying FL for AI-RAN enabled MEC system faces a critical challenge, i.e., statistical (data) heterogeneity \cite{Y_Cheng_Pre_clustering_CFL_Data_System}, a.k.a. non-independent and identically distributed (non-IID) \cite{C_Li_FL_Soft_Clustering}. As per \cite{X_Li_FL_Adaptive_Weights_Non_IID}, it is inevitable to present the non-IID data across the edge devices due to various factors (e.g., different environments, distinct data sources, and various hardware systems). Because of the presence of non-IID data, FL encounters the performance degradation phenomenon when learning a shared global model \cite{L_Zou_Towards_Satellite_Non_IID_Imagery}, which may hinder FL's practical application. To be specific, due to the significant data distribution gap for the local clients, after training, the local updates will diverge unfavorably \cite{Z_Li_Feature_Matching_Data_Synthesis}. This will lead to a misled global model and degrade the accuracy \cite{Z_Li_Feature_Matching_Data_Synthesis}. Therefore, it is imperative to tackle the non-IID data issue faced by FL.
	
	\par
	To circumvent the non-IID data issue, for AI-RAN enabled MEC system, we propose a \underline{m}ulti-\underline{p}rototype-guided \underline{fed}erated \underline{k}nowledge \underline{d}istillation (MP-FedKD) approach in this work. It includes: self-knowledge distillation (SKD) technique, a proposed conditional hierarchical agglomerative clustering (CHAC) for generating multiple prototypes of each class, a novel prototype alignment scheme, and a novel loss function (designed based on COREL \cite{K_Kenyon_Clustering_Representation} loss). The \textit{\textbf{motivations}} for designing the proposed method are summarized as below:
	
	\begin{itemize}
		\item To deal with the non-IID data issue, one promising method is to employ knowledge distillation (KD), as integrating KD into FL to handle data heterogeneity is an effective manner \cite{J_Tang_FedRAD}. Concretely, KD is a technique that is capable of benefiting the smaller student network's training process by the direction of a larger teacher network (TN) \cite{L_Zou_Cyber_Attacks_Prevention_Prosumer_EV}. Particularly, in KD, the TN's knowledge is used to guide the student network (SN) training. \textit{Nevertheless, although KD allows the larger network (i.e., teacher network) to be leveraged in a condensed way, the inference on the TN will become a burden for KD's practical application \cite{M_Ji_Refine_Myself_Teaching_Myself}. Apart from this, to prepare a proper teacher network that can correspond to a target SN is difficult \cite{H_Lee_self_knowledge_distillation}.} To overcome those drawbacks of KD, a technique called \textbf{self-knowledge distillation} can be taken into consideration, as there is no prior necessity to pre-train a teacher network in advance for SKD as per \cite{M_Ji_Refine_Myself_Teaching_Myself}. To be more specific, in SKD, the TN is the SN itself \cite{H_Lee_self_knowledge_distillation}. Accordingly, in this work, instead of conventional KD, we employ SKD as one of the techniques to construct the proposed solution.
		\item Except from KD, on the basis of \cite{Y_Tan_FedProto}, prototype-based mechanism can also be considered applying into FL for the clients with heterogeneous setting. Many works designed the FL-based solution based on a single prototype-based strategy, such as \cite{Y_Tan_FedProto, X_Mu_FedProc, B_Yan_FedRFQ, Y_Tan_FL_Pre_Trained_Contrastive, G_Yan_FedVCK_Non_IID_Condensed} (details can be found in Section \ref{sec_2B_single_prototype}). \textit{Nevertheless, to generate local prototypes, average operation is adopted by \cite{Y_Tan_FedProto, X_Mu_FedProc, B_Yan_FedRFQ, Y_Tan_FL_Pre_Trained_Contrastive, G_Yan_FedVCK_Non_IID_Condensed}, which will cause the loss of some useful information \cite{B_Zhang_Self_Guided_Cross_Guided_Learning}.} To let the model get a more comprehensive grasp of the samples' feature information, multi-prototype generation can be delved into as per \cite{J_Zhang_Dual_Expert_Distillation}. Consequently, in this work, we employ the multi-prototype-based strategy (MPS) to devise the proposed solution, where we design a \textbf{conditional hierarchical agglomerative clustering (CHAC) approach} based on the hierarchical agglomerative clustering (HAC) method \cite{L_Zou_Imbalance_Cost_Energy_Scheduling} (details will be illustrated in Section \ref{Sec_Methodology_HAC_Multi_Prototype}). The reason for we adopt HAC is that compared to the non-hierarchical method, because of the adopted dendrogram, the hierarchical approach can provide more information \cite{H_R_A_Putri_segmentation_hierarchical_clustering}.
		\item Global prototype per class can be obtained by aggregating the local prototypes of all participating clients belonging to the same category via average manner \cite{Y_Tan_FedProto}. \textit{As mentioned before, such a manner will result in some useful information being lost \cite{B_Zhang_Self_Guided_Cross_Guided_Learning}.} To mitigate this phenomenon, we consider making the global prototypes learn from the historical local embedding. For simplicity, we call this manner as \textbf{prototype alignment} in this work. To the best of our knowledge, the detailed process of PA in this article is novel.
		\item A basic insight of COREL loss is that samples should exhibit similarity to other samples that belong to the same class and be distinct from samples in different categories \cite{K_Kenyon_Clustering_Representation}. Inspired by this insight, we attempt to design \textbf{a novel loss function}, in which we consider aligning the \underline{l}ocal \underline{em}bedding with a \underline{g}lobal \underline{p}rototype of the same class, while separating the local embedding with the global prototypes of other classes. For simplicity, we refer to this loss as \textbf{LEMGP loss}. Details will be provided in the \textit{contribution} part.
	\end{itemize}
	\textit{\textbf{Motivated by the above facts, the major contributions}} of this article are highlighted as follows:
	
	\begin{itemize}
		\item In order to deal with the non-IID data issue in AI-RAN enabled MEC system, we propose a \underline{m}ulti-\underline{p}rototype-guided \underline{fed}erated \underline{k}nowledge \underline{d}istillation (MP-FedKD) approach. This method involves SKD technique, CHAC for multi-prototype generation, a new prototype alignment mechanism, and the newly designed LEMGP loss.
		\item For SKD, we regard the previous round of local model as teacher model \cite{L_Zou_Towards_Satellite_Non_IID_Imagery}, where its knowledge will be distilled to guide the current local model training. This manner can overcome the shortcomings of conventional KD, as mentioned above, in handling the non-IID data.
		\item With respect to the multi-prototype generation per client, to produce the multi-prototypes per class, we propose a conditional hierarchical agglomerative clustering (CHAC) method, which is constructed based on HAC approach \cite{L_Zou_Imbalance_Cost_Energy_Scheduling}. In particular, the embedding of each sample (i.e., the output of the representation layers) belonging to the same class is regarded as the input. Moreover, motivated by \cite{L_Zou_Imbalance_Cost_Energy_Scheduling}, the sum of square (SSQ)-based Ward’s approach is used by CHAC to decide clusters merging. 
		\item In \cite{X_Yang_FedAS}, the embedding obtained via global model in the current round is considered to learn from the local embedding obtained by using the previous round local model. In such a way, as per \cite{X_Yang_FedAS}, the local representation knowledge can be aligned with the shared parameter. Inspired by this, for the sake of improving the situation of useful information loss caused by average manner, we consider incorporating the local embedding with the global prototypes. Particularly, motivated by  \cite{X_Yang_FedAS}, we let the global prototype per class in current round to learn from local embedding vectors of the corresponding class that are generated by utilizing the the previous round local model (called prototype alignment in this work for convenience).

		\item We propose a novel loss (named LEMGP loss) by referring to \cite{K_Kenyon_Clustering_Representation} and \cite{J_Deuschel_Multi_Prototype_Few_shot}. For LEMGP loss, it includes two parts: attractive part and repulsive part. For the attractive part, we propose a weighted MSE-loss, in which local embeddings and global prototype of the same class are used. For the repulsive part, it involves logarithmic function, exponential function, and weighted MSE-loss, where the similarity between the local embedding and the global prototypes is considered. 
		\item Evaluation experiments are conducted based on six datasets, \textit{CIFAR-10} \cite{A_Krizhevsky_Learning_multiple_tiny}, \textit{MNIST}, \textit{Fashion-MNIST} \cite{H_Xiao_Fashion_MNIST}, \textit{EuroSAT} \cite{P_Helber_EuroSAT}, \textit{M+F} and \textit{C+E}, under various non-IID settings. M+F dataset is the combination of MNIST and Fashion-MNIST, while C+E is the combination of CIFAR-10 and EuroSAT. The results clarify that the proposed MP-FedKD is more competitive, compared to the baselines. For instance, the accuracy improvement by the proposed method lies between $1.98\%\sim28.70\%$ for EuroSAT dataset when considering the number of clients as $10$. The improvement can also be recognized from other settings (e.g., the number of clusters, various datasets). 
				
	\end{itemize}

\vspace{-0.1cm}
	\section{Related Work}
	\label{Sec2_Related_Work}
	As the proposed MP-FedKD is designed based on KD and prototypes, in this section, we will list several previous literature in KD-based FL, single prototype-based strategy, and multi-prototype-based strategy. Besides, we also point out the difference between the given previous literature and this work.

\vspace{-0.1cm}
	\subsection{KD-based FL}
    Some previous literature has taken into account employing KD into FL such as \cite{ L_Zou_Cyber_Attacks_Prevention_Prosumer_EV,  A_H_Mohamed_Combining_Client_Selection_KD, S_Ge_FedAMKD_Adaptive_Mutual_KD_FL, D_Yao_FedGKD, Y_Sahraoui_FedRx_Distillation, T_D_Nguyen_HFL_MEC_KD, J_Chai_PFRL_AAV_MEC, G_Sun_FeDistSlice_Federated_Policy, T_Kwantwi_PFL_MEC_RAN, Z_Wang_Communication_Efficient_PFL_DT, G_Liao_PFL_SKD_VEC}. Specifically, in \cite{L_Zou_Cyber_Attacks_Prevention_Prosumer_EV}, \texttt{E-FPKD} was proposed for cyber-attack prevention, where each client is formed by a teacher network and a student network. In \cite{A_H_Mohamed_Combining_Client_Selection_KD}, \texttt{FedCCSKD} was proposed, where the global model is regarded as the teacher model and the local client model is treated as the student model. In \cite{S_Ge_FedAMKD_Adaptive_Mutual_KD_FL}, \texttt{FedAMKD} was proposed to attend to the data heterogeneity issue. To be specific, in \texttt{FedAMKD}, an adaptive mutual KD is deployed on the local client to transfer the knowledge between the local model and the global model. However, \cite{L_Zou_Cyber_Attacks_Prevention_Prosumer_EV, A_H_Mohamed_Combining_Client_Selection_KD, S_Ge_FedAMKD_Adaptive_Mutual_KD_FL} do not consider self-knowledge distillation (SKD) technique, which is one of the concerns in this article. \cite{D_Yao_FedGKD} proposed two methods named \texttt{FEDGKD} and \texttt{FEDGKD-VOTE}, which can be adopted to address non-IID data issue. For \texttt{FEDGKD}, an ensemble model is generated by aggregating the historical global models (HGMs) in the server, which will be sent to the clients to perform as a teacher to guide local clients' training. Different from \texttt{FEDGKD}, \texttt{FEDGKD-VOTE} does not adopt aggregation operation to produce the ensemble model. The HGMs in \texttt{FEDGKD-VOTE} will be sent to local clients directly to instruct the local training. In contrast with \texttt{FEDGKD}, ensemble model is not adopted by this work. Compared to \texttt{FEDGKD-VOTE}, instead of using HGMs, we leverage previous round local model to guide local model training. In \cite{Y_Sahraoui_FedRx_Distillation}, KD-based FL approach has been designed to sense disease in mobile edge computing-based network. In \cite{T_D_Nguyen_HFL_MEC_KD}, dual KD is proposed to combine with FL to devise a hierarchical FL-based approach in mobile edge computing networks, in which the global model and an ensemble regional model are used to guide local models' training. In \cite{J_Chai_PFRL_AAV_MEC}, multi-AAV (autonomous
aerial vehicle) assisted mobile edge computing network was considered. In such a network, \texttt{FKD-PDQN} was proposed to mitigate the communication cost, where KD is leveraged. It is noteworthy that Multi-access Edge Computing is formerly known as Mobile Edge Computing as per \cite{W_Fan_Joint_Task_Offloading_MEC}. Although \cite{Y_Sahraoui_FedRx_Distillation, T_D_Nguyen_HFL_MEC_KD, J_Chai_PFRL_AAV_MEC} all took into account the Mobile Edge Computing  network, AI-RAN, SKD and prototype-based strategy do not be included in those works, which are important to this work. In \cite{G_Sun_FeDistSlice_Federated_Policy}, \texttt{FeDistSlice} was proposed to investigate the heterogeneity issues in mobile edge computing-based Radio Access Network (RAN) slicing, where KD and FL are adopted. In \cite{T_Kwantwi_PFL_MEC_RAN}, \texttt{PerFedD3QL} was presented to make SCTOSRA decisions towards MEC-enabled RAN, where FL, KD and dueling double deep Q-learning are involved. However, SKD and multi-prototypes-based strategy do not considered by \cite{G_Sun_FeDistSlice_Federated_Policy} and \cite{T_Kwantwi_PFL_MEC_RAN}, which are two significant parts of this work. In \cite{Z_Wang_Communication_Efficient_PFL_DT} and \cite{G_Liao_PFL_SKD_VEC}, SKD had been adopted to integrate with FL to design the personalized FL (pFL)-based approach. However, compared with those two works, we do not consider personalized FL scheme in this paper. To be more specific, we consider both model aggregation and multi-prototype aggregation to separately generate global model and global prototypes. In addition to this, the loss function designed in this work is different from the one proposed in \cite{Z_Wang_Communication_Efficient_PFL_DT} and \cite{G_Liao_PFL_SKD_VEC}.

\vspace{-0.1cm}	
    \subsection{Single Prototype-based Strategy}
	\label{sec_2B_single_prototype}
    Single prototype-based strategy has been applied by several works to build the FL-based method, such as \cite{Y_Tan_FedProto, X_Mu_FedProc, B_Yan_FedRFQ, Y_Tan_FL_Pre_Trained_Contrastive, G_Yan_FedVCK_Non_IID_Condensed, T_Gao_FedPC, L_Chai_Prototype_based_fine_tuning_mitigating, Y_Zhou_FedSA, T_K_Tran_FedNTProto, H_Li_FedCPG}. \textit{It is worth noting that a single prototype-based strategy in this work means that only one prototype will be generated for one class.} In \cite{Y_Tan_FedProto}, a federated prototype learning framework dubbed \texttt{FedProto} was presented, where the local prototype of a class is a mean vector obtained by averaging the embedding vectors of all data belonging to that class. In \cite{X_Mu_FedProc}, an approach called \texttt{FedProc} was proposed towards non-IID data, in which the local prototype per class is obtained by calculating the mean vector of all data samples' feature vectors of that class (same as \texttt{FedProto}). In \cite{B_Yan_FedRFQ}, \texttt{FedRFQ} was presented, which has the ability of addressing the non-IID data issue. To be specific, for \texttt{FedRFQ}, a pooling method named \textit{SoftPool} is introduced to get the averaged pooled prototypes based on each class, where the averaged pooled prototypes will be sent to the server to obtain the global prototypes. In \cite{Y_Tan_FL_Pre_Trained_Contrastive}, \texttt{FedPCL} was proposed. Specifically, in \texttt{FedPCL}, local prototype of each class is computed by the mean of the output of the projection network. The local prototypes of all clients in \texttt{FedPCL} will be aggregated by the server to generate the global prototypes. Then contrastive learning will be conducted based on the obtained prototypes. In \cite{G_Yan_FedVCK_Non_IID_Condensed}, \texttt{FedVCK} was proposed to deal with the Non-IID data issue, where knowledge condensation is occurred in each client to generate small dataset. Over the condensed knowledge dataset, a single local logit prototype of each class will be calculated. In \cite{T_Gao_FedPC}, \texttt{FedPC} was presented, where Non-IID data issue is considered. To be specific, in each local client, a single local prototype corresponds to one class will be computed, and then a generalized prototype (average value of all the local prototypes) will be calculated. Both local prototypes and the generalized prototype will be used by the server to build  feature matrix. In \cite{L_Chai_Prototype_based_fine_tuning_mitigating}, the local prototype of a class at current round is generated by using the global feature extractor of previous round on the local dataset, which will be used to fine-tune the the global classifier of previous round. As for \cite{Y_Zhou_FedSA, T_K_Tran_FedNTProto, H_Li_FedCPG}, they followed \texttt{FedProto} to generate the local prototypes. Different from \cite{Y_Tan_FedProto, X_Mu_FedProc, B_Yan_FedRFQ, Y_Tan_FL_Pre_Trained_Contrastive, G_Yan_FedVCK_Non_IID_Condensed, T_Gao_FedPC, L_Chai_Prototype_based_fine_tuning_mitigating, Y_Zhou_FedSA, T_K_Tran_FedNTProto, H_Li_FedCPG}, in this work, conditional hierarchical agglomerative clustering \cite{L_Zou_Imbalance_Cost_Energy_Scheduling} (HAC) is presented to obtain the multi-prototypes for each class in each local client, which will be used to form the prototype alignment scheme and the LEMGP loss. 
    	
    \subsection{Multi-Prototype-based Strategy}
    \par	
    Multi-prototype-based strategy have been explored by several studies such as \cite{J_Deuschel_Multi_Prototype_Few_shot, L_Wang_Taming_Cross,   X_Xu_Multiple_Adaptive_Prototypes_PFL, Y_Bi_Multi_Prototype_Embedding_Refinement, M_Le_FedMP_Multi_Prototype, K_Zhang_Federated_Learning_Heterogeneous_GDBD, S_F_Peng_Multi_Granularity_Aggregation_Network, R_Fan_NAPG}. In \cite{J_Deuschel_Multi_Prototype_Few_shot}, k-means was adopted to generate multi-prototypes for each class based on corresponding support set's embedding, where the cluster centers are regarded as prototypes that will be used to design COREL loss. However, FL scheme does not considered by \cite{J_Deuschel_Multi_Prototype_Few_shot}, whereas the proposed method of this work covers FL scheme. Besides, though COREL loss is studied by \cite{J_Deuschel_Multi_Prototype_Few_shot}, it is different from this work. Concretely, the proposed LEMGP loss of this work is created by using the global prototypes. In \cite{L_Wang_Taming_Cross}, \texttt{FedPLVM} was proposed, in which multi-prototypes are considered to be generated in both the local clients side and the server side via FINCH algorithm \cite{S_Sarfraz_FINCH}. Nevertheless, although \cite{L_Wang_Taming_Cross} integrates the multi-prototype-based strategy into FL, they do not consider KD, prototype alignment and COREL loss \cite{K_Kenyon_Clustering_Representation}, whereas those three techniques are the significant points of this paper. In \cite{X_Xu_Multiple_Adaptive_Prototypes_PFL},  \texttt{mapFL} (a personalized FL approach) was proposed, in which the multi-prototypes of a category for a client are retrieved from the local memory network of that client. In \cite{Y_Bi_Multi_Prototype_Embedding_Refinement}, a momentum-based strategy was proposed to obtain the prototype, rather than directly using the cluster centers as the prototypes. In \cite{M_Le_FedMP_Multi_Prototype} and \cite{K_Zhang_Federated_Learning_Heterogeneous_GDBD}, k-means and Kmeans++ were used to generate multi-prototypes of each class for each client, respectively. In \cite{S_F_Peng_Multi_Granularity_Aggregation_Network}, a module called AMPA was proposed, in which multi-prototypes (e.g., support prototype, query prototype, mainstay prototype) were involved. In \cite{R_Fan_NAPG}, k-means clustering algorithm was also used to generate multi-prototypes for each class, where the input is the normalized deep features of each class. Different from \cite{X_Xu_Multiple_Adaptive_Prototypes_PFL, Y_Bi_Multi_Prototype_Embedding_Refinement, M_Le_FedMP_Multi_Prototype, K_Zhang_Federated_Learning_Heterogeneous_GDBD,  S_F_Peng_Multi_Granularity_Aggregation_Network, R_Fan_NAPG}, in this paper, we attempt to design a method (called CHAC) based on the HAC approach to produce multi-prototypes for each class.

    \vspace{-0.1cm}

    \section{System Model}
     In this section, we will go into the system model of the considered AI-RAN Enabled MEC system, which contains the AI-RAN Enabled MEC network model, the FL model and the prototype-based model.  
    

    \subsection{Model of AI-RAN Enabled MEC System}
\begin{figure} [t!]
    	\centering
    	\includegraphics[scale = 0.47]{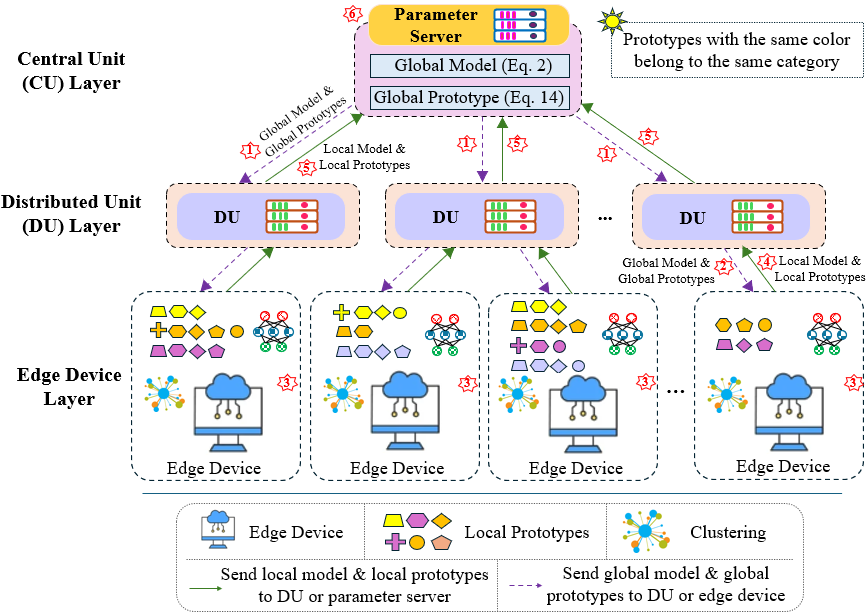}
    	\captionsetup{font=small}
    	\caption{System model of the AI-RAN enabled MEC system, where each edge device can generate local prototypes and local model through training.}
    	\label{fig_System_Model}
    	\vspace{-0.4cm}
    \end{figure}

    Fig. \textcolor{red}{\ref{fig_System_Model}} shows the system model of the considered AI-RAN enabled MEC system, which contains three layers: central unit (CU) layer, distributed unit (DU) layer, and edge device layer. Concretely, the CU layer consists of a parameter server (i.e., central server) that will be used for model aggregation in FL process. As for the DU layer, it is formed by a set of distributed units (DUs), denoted as $\mathcal{G}=\{1, 2, ..., G\}$. Inspired by \cite{A_Younis_Communication_Efficient_NG_RAN}, we assume each DU associates with multiple edge devices. The set of edge devices under the coverage of a DU $g\in\mathcal{G}$ is indexed by $\mathcal{M}_g=\{1, 2, ..., M_g\}$. It is noteworthy that we assume that each edge device can at most associate with one DU for simplicity in this work. Thus, the edge devices across the entire system can be given as $\mathcal{M}=\mathcal{M}_1 \cup \mathcal{M}_2 \cup...\mathcal{M}_G$. For each edge device $m$, we consider it has its own dataset, which is denoted as $\mathcal{D}_m=\{(x_m^\textrm{1}, y_m^\textrm{1}), (x_m^\textrm{2}, y_m^\textrm{2}), ..., (x_m^\textrm{n}, y_m^\textrm{n})\}$. Here, $x_m^\textrm{n} \in \mathcal{X}_m$ and $y_m^\textrm{n} \in \mathcal{Y}_m$ represent the $n^\textrm{th}$ sample feature in the set of $\mathcal{X}_m=\{x_m^\textrm{1}, x_m^\textrm{2}, ..., x_m^\textrm{n}\}$ and the corresponding $n^\textrm{th}$ label in the set of $\mathcal{Y}_m=\{y_m^\textrm{1}, y_m^\textrm{2}, ..., y_m^\textrm{n}\}$, respectively. Then, the overall dataset across all edge devices can be represented by $\mathcal{D}=\cup_{m=1}^{|\mathcal{M}|} \mathcal{D}_m$ with $|\mathcal{D}|$ samples, where $|\mathcal{M}|$ denotes the number of all edge devices of the entire system. 

    \par
   Moreover, for the considered AI-RAN enabled MEC system, its components (i.e., the central server, multiple DUs and multiple edge devices) are used to construct the FL scheme. Specifically, each edge device is treated as each client in FL, on which we consider performing both a deep learning model and the clustering to generate the local model and the local prototypes (illustrated in Section \ref{prototype_based_model_MEC_Nework}), respectively. As per \cite{A_Younis_Communication_Efficient_NG_RAN}, each DU can be leveraged to relay the models between end-users and the server located in the CU layer. Therefore, in this work, regarding DUs, we retain their relay function to enable the model and prototype exchange between the edge devices and the central server located in the CU layer. With respect to the central server, it plays the role of producing the global model in each round (explained in Section \ref{FL_MEC_network}). Further, as per \cite{Y_Sahraoui_FedRx_Distillation}, data transmission is able to be performed by leveraging the robust wireless communication technologies (RWCT). Thus, for simplicity, the data transmission (i.e, model and prototype exchange) will be fulfilled by using RWCT in this work. 
   \par
   In the case of the training process in the considered system, we consider it as below: 1) in the beginning, the parameter server will distribute the initialized global model \& global prototypes to each DU. 2) After receiving the global model \& global prototypes, each DU will relay them to its covered edge devices, respectively. 3) Then each edge device will perform the local training to generate the local model and local prototypes. 4) Afterwards, the produced local model and local prototypes will be uploaded to DUs. 5) Each DU will execute the relay process to disseminate the local models and local prototypes for its covered edge devices to the parameter server. 6) With the received local models and local prototypes, global model aggregation and global prototypes aggregation processes will be conducted. These steps are performed iteratively. Notably, as each edge device is assumed to be connected with only one DU, the local model and local prototypes belonging to the same client will not be duplicated on the central server.
   \par
   Next, we will elaborate on the relevant content regarding the FL model and prototype-based model in the considered AI-RAN Enabled MEC system.

    

	\subsection{FL Model in AI-RAN Enabled MEC System}
    \label{FL_MEC_network}
    In this part, we will give a detailed explanation of the employed FL model in the considered system. Let $\omega$ indicate the global model, the objective of general FL (i.e., FedAvg \cite{H_B_McMahan_Communication_Decentralized}) can be defined as follows \cite{L_Zou_EFCKD}:
	\begin{subequations}\label{Opt_1}
		\setlength{\abovedisplayskip}{3.2pt}
		\setlength{\belowdisplayskip}{3.2pt}
		\begin{align}
			\underset{\boldsymbol{\omega}} \min 
			&\; \quad f(\omega) = \sum_{m=1}^{\left | \mathcal{M}_c \right |} \frac{\left| \mathcal{D}_m \right|}{\left| \mathcal{D} \right|}F_m(\omega),  \tag{1} 
		\end{align}	
	\end{subequations}
	where $\mathcal{M}_c$ denotes a set of participating clients that owns data of class $c$, while $|\mathcal{M}_c|$ is the number of those clients. Notably, $\mathcal{M}_c \leq \mathcal{M}$, this is because in practice, as noted in \cite{L_Zou_Cyber_Attacks_Prevention_Prosumer_EV}, it is not feasible to assume all clients are always able to join training. $F_m(\omega)= \frac{1}{\left| \mathcal{X}_m \right|}\Sigma{_{n=1}^{|\mathcal{X}_m|}}f_m^\textrm{n}(\omega)$, here, $|\mathcal{X}_m|$ represents the number of the samples belonging to the edge device $m$, while $n$ denotes the $n^\textrm{th}$ data sample in the set of $\mathcal{X}_m$. As for $f_m^\textrm{n}(\omega)$, it is used to indicate the user-specified loss function with the given input (each edge device's $n^\textrm{th}$ data sample). Let $\omega_{m}$ denote the local model of client $m$, consider the general FL, we can compute the $\omega$ as below \cite{L_Zou_Cyber_Attacks_Prevention_Prosumer_EV}:
	\begin{subequations}\label{Opt_2}
		\setlength{\abovedisplayskip}{3.2pt}
		\setlength{\belowdisplayskip}{3.2pt}
		\begin{align}
			\omega = \sum_{m=1}^{\left |\mathcal{M}_c \right |} \frac{\left| \mathcal{D}_{m} \right|}{\left| \mathcal{D} \right|}\omega_{m},  \tag{2} 
		\end{align}	
	\end{subequations}
	Through \eqref{Opt_2}, it can be seen that the parameter-level fusion \cite{Z_Zhu_ISFL_FL_Non_IID_Importance} is employed by the general FL. Thus, the global performance may depend on the local training. 
    \par
    As per \cite{Z_Zhu_ISFL_FL_Non_IID_Importance}, each client has a tendency to update its local model to a local optimum which is suitable for its own local training data. Consequently, the non-IID data phenomenon across all the clients may pose a challenge to FL, which still needs to be further improved. As aforementioned, to cope with the heterogeneous issue, we use the prototype-based mechanism (PBM). Thus, in the following, we will provide an in-depth discussion regarding the prototype-based model in the considered AI-RAN Enabled MEC system.

    
	\subsection{Prototype-based Model in AI-RAN Enabled MEC System}
    \label{prototype_based_model_MEC_Nework}
	
	\par
	In this subsection, we will take a close look at the consided prototype-based model in the considered system.
	\par
	Before giving the definition of the prototype, it is necessary to refer to the structure of the deep learning (DL)-based model placed on each edge device in the considered system. Generally, the DL model consists of two parts: 1) the representation layers \cite{Y_Tan_FedProto} \big(denoted as $\Upsilon(\cdot)$\big), and 2) classifier \cite{L_Zou_Cyber_Attacks_Prevention_Prosumer_EV} \big(denoted as $\Omega(\cdot)$\big). Considering the $C$-classification scenario, for which we can denote a set of categories in the central server side as $\mathcal{C}=\{1, 2, ..., C\}$. As for each client, since we consider non-IID setting, a set of categories owned by each client $m$ can be represented as $\mathcal{C}_m=\{1, 2, ..., C_m\}$. 

    \begin{figure*} [t!]
		\centering
		\includegraphics[scale = .7]{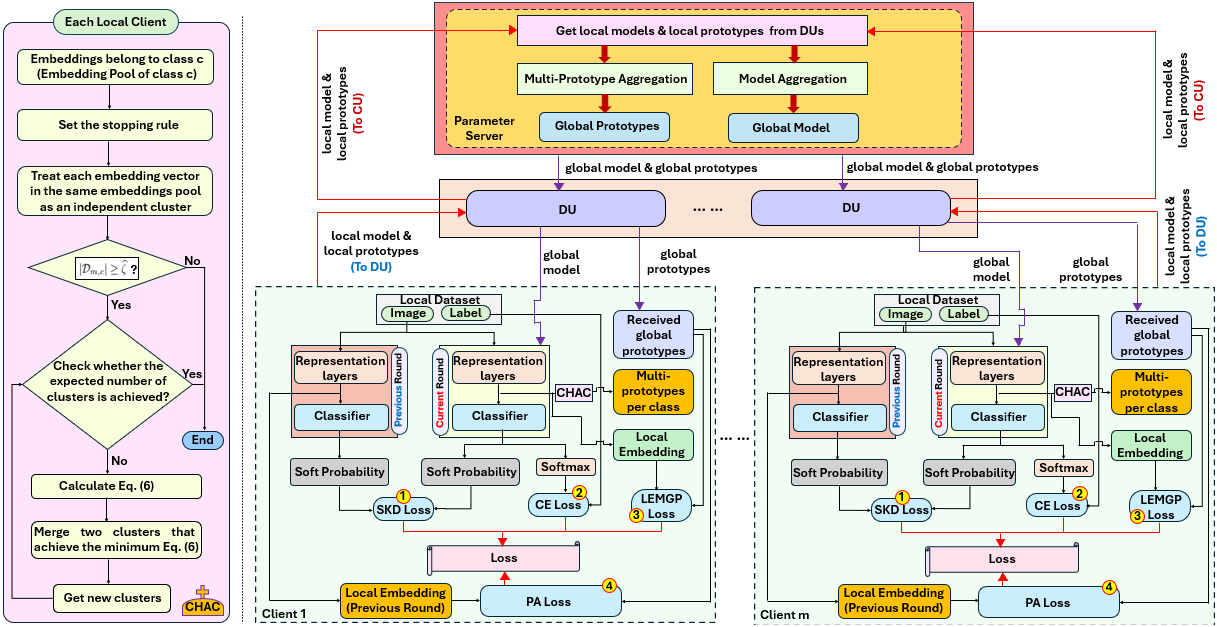}
		\vspace{-0.2cm}
		\caption{CHAC for per-class multi-prototype generation on the client (left) and architecture of the proposed MP-FedKD (right).}
		\label{fig_Solution} 
		\vspace{-0.4cm}
	\end{figure*}
    
	\par
	\textbf{Local Prototypes} \quad For each \textit{local prototype} of client $m$, originally, based on \cite{Y_Tan_FedProto}, it can be obtained by averaging the embedding vectors of the instances belonging to the category $c \in \mathcal{C}_m$, where each embedding vector is the output generated by representation layers. It is worth noting that the prototype obtained in this way corresponds to one class. Namely, a single prototype-based strategy is adopted by \cite{Y_Tan_FedProto}. However, as mentioned above, owing to the average manner, adopting a single prototype-based strategy will lead to some loss of useful information \cite{B_Zhang_Self_Guided_Cross_Guided_Learning}. Thus, instead, in this work, we borrow the concept of multi-prototype for a single class from \cite{J_Deuschel_Multi_Prototype_Few_shot} to compensate for the drawback of the single prototype-based strategy, where the multiple local prototypes for a single class is defined by the cluster centroids of the data embedding belonging to the same category. To this end, clustering technique deployed on each local edge device is taken into consideration to generate the local prototypes for each class in this article. For simplicity, let $P_{m, c}^{i}$ indicate $i^\textrm{th}$ local prototype of class $c$ for edge device $m$, we give the following mathematical representation to compute the multiple local prototypes for class $c$:
	\begin{subequations}\label{Opt_3}
		\setlength{\abovedisplayskip}{3.2pt}
		\setlength{\belowdisplayskip}{3.2pt}
		\begin{align}
			\{P_{m, c}^i\}_{i=1}^{|\mathcal{P}_{m, c}|} \xleftarrow{\textrm{cluster centroids}}	cluster(\{\Upsilon(x_m^\textrm{n};y_m^\textrm{n}=c)\}_{n=1}^{\left| \mathcal{D}_{m, c} \right|}),  \tag{3} 
		\end{align}	
	\end{subequations}
	where $\mathcal{P}_{m, c} = \{1, 2, ..., P_{m, c}\}$ denote a set of local prototypes per class $c$ for client $m$, and $|\mathcal{P}_{m, c}|$ is used to indicate the number of local prototypes belonging to the same class $c$ in client $m$. $\mathcal{D}_{m, c}$ denotes the dataset that belongs to class $c$ for client $m$, which is the subset of $\mathcal{D}_m$. 
	
	\par
	\textbf{Global Prototype} \quad Regarding the global prototype, in \cite{Y_Tan_FedProto}, it is obtained by aggregating the local prototypes of all participating clients belonging to the same class in the server, where one local prototype corresponds to one class in each client in \cite{Y_Tan_FedProto}. Same manner was also considered by \cite{L_Wang_Taming_Cross}, where multiple local prototypes belonging to the same class lie at one client is taken into account.  For simplicity, we follow both \cite{Y_Tan_FedProto} and \cite{L_Wang_Taming_Cross} to produce each global prototype. Specifically, we aggregate the multiple local prototypes of the same class per participating client in the central server to produce the global prototype of that class (denoted as $\overline{P_{c}}$). The mathematical representation is given as below:
	\begin{subequations}\label{Opt_4}
		\setlength{\abovedisplayskip}{3.2pt}
		\setlength{\belowdisplayskip}{3.2pt}
		\vspace{-0.06cm}
		\begin{align}
			\overline{P_{c}} = \frac{1}{|\mathcal{M}_c||\mathcal{P}_{m, c}|}\sum_{m=1}^{\left |\mathcal{M}_c\right |}\sum_{i=1}^{\left | \mathcal{P}_{m, c}\right |} \frac{\left| \mathcal{D}_{m, c} \right|}{\left| \mathcal{D}_c \right|}P_{m, c}^{i},  \tag{4} 
		\end{align}	
	\end{subequations}
	where $\mathcal{D}_c$ indicates the dataset of all the participating clients that owns class $c$'s data samples, which can be defined as $\mathcal{D}_c=\bigcup_{m=1}^{|\mathcal{M}_c|}\mathcal{D}_{m, c}$. $|\mathcal{D}_c|$ is the amount of data in $\mathcal{D}_c$. More elaboration will be provided in the upcoming section.


     \vspace{-0.2cm}
	\section{Methodology}
	\label{section_4_methodology}
	\par
	In this section, we will introduce the proposed MP-FedKD approach to process the non-IID data in the considered AI-RAN enabled MEC system, as shown in Fig. \textcolor{red}{\ref{fig_Solution}}. Specifically, the proposed MP-FedKD approach consists of: 1) self-knowledge distillation, 2) multi-prototype generation via CHAC, 3) prototype alignment, 4) LEMGP loss, 5) multi-prototype and model aggregation. Next, we will go into detail.

\vspace{-0.2cm}
	\subsection{Self-Knowledge Distillation}
	As stated above, SKD is leveraged to solve the non-IID data issue in this paper. For SKD, as per \cite{H_Jin_pFedSD}, when updating the local client, the knowledge of latest updated model is able to be distilled to guide the training process of the local client. To this end, in this work, the local model generated in the previous round (i.e., round $t-1$), $\omega_m^{t-1}$, will be treated as the teacher model, which will be leveraged to teach the training of the current round (i.e., round $t$) local model (i.e., student model), denoted as $\omega_m^{t}$. Note that SKD is performed from the second round \cite{L_Zou_Towards_Satellite_Non_IID_Imagery} since it is necessary to prepare the latest historical local model. SKD loss on edge device $m$ can be defined as below \cite{B_Peng_Correlation_Congruence_ICCV}:
	\begin{subequations}\label{Opt_5}
		\setlength{\abovedisplayskip}{3pt}
		\setlength{\belowdisplayskip}{3pt}
		\begin{align}
			&\label{Opt_5:const1}
			\mathcal{L}_m^\textrm{SKD} =  \frac{1}{|\mathcal{D}_m|}\sum_{n=1}^{|\mathcal{D}_m|}\tau^2KL\Big(\xi_m^c(x_m^\textrm{n}), \widehat{\xi_m^c}(x_m^\textrm{n})\Big), \tag {5}
		\end{align}	
	\end{subequations}
	where $KL(\cdot, \cdot)$ represents the Kullback-Leibler (KL) divergence, while $\tau$ indicates the temperature. $\xi_m^c(x_m^\textrm{n})$ and $\widehat{\xi_m^c}(x_m^\textrm{n})$ separately denote the soft probabilities of teacher model \& student model, which can be computed by \textit{Softmax with Temperature scaling} \cite{G_Hinton_Distilling_Knowledge_NN}. 
	\par
	Note that the student model will be utilized to produce multiple local prototypes of each class on each local edge device in the considered system, which will be explained next.

    \vspace{-0.1cm}
	\subsection{CHAC-based Multi-Prototype Generation}
	\label{Sec_Methodology_HAC_Multi_Prototype}
	To generate multi-prototype of each class on the local side, a clustering technique is designed based on the hierarchical agglomerative clustering (HAC) approach, as more information can be offered by the hierarchical approach than the non-hierarchical method \cite{H_R_A_Putri_segmentation_hierarchical_clustering}. To be specific, to adapt to this work, we add an \textit{additional condition} to the original HAC approach. For convenience, we call the devised approach as conditional HAC (CHAC) \big(Fig. \textcolor{red}{\ref{fig_Solution}} (left)\big).
      
    \par
    Concretely, we conduct CHAC approach on each local edge device, where the local embeddings (the output of representation layers) of each class $c$ are utilized as the input. Specifically, we denote the embedding of the data sample $(x_m^\textrm{n}, y_m^\textrm{n}) \in \mathcal{D}_{m, c}$ as $\varrho_{m, c}^\textrm{n}$, and it is given as $\varrho_{m, c}^\textrm{n} = \Upsilon(x_m^\textrm{n};y_m^\textrm{n}=c)$. In the beginning, each $\varrho_{m, c}^\textrm{n}$ is regarded as a single cluster. Those clusters build a nested cluster set via the dendrogram \cite{L_Zou_Imbalance_Cost_Energy_Scheduling}. Then we will merge the nearest pair of clusters until reaching the stopping rule. 
    \par
    To determine which clusters can be merged, the \textit{sum of square (SSQ)-based Ward's method} \cite{S_Hirano_Comparison_clustering} is employed. Consider two clusters belonging to the same class, denoted as $B_1$ and $B_2$. The $B_1$ and $B_2$ are considered containing $v_1$ and $v_2$ examples, respectively. Normally, the embedding vector of a data sample is called an example. Let $Q$ denote the overall number of elements in an example $h$, where an element is a value in the embedding vector. Let $E_{B_{1}}^{h, q}$ and $E_{B_{2}}^{h, q}$ denote the $q^\textrm{th}$ element of each example $h$ in cluster $B_1$ and $B_2$, respectively,  we can compute $\bigtriangleup SSQ_{{B_{1}}{B_{2}}}$ as \cite{S_Hirano_Comparison_clustering}:
	\begin{subequations}\label{Opt_6}
		\setlength{\abovedisplayskip}{3pt}
		\setlength{\belowdisplayskip}{3pt}
		\vspace{-0.1cm}
		\begin{align}
			&\label{Opt_6:const1}
			\bigtriangleup SSQ_{{B_{1}}{B_{2}}} = \frac{v_{1}v_{2}}{v_{1}+v_{2}}\sum_{q=1}^{Q} (\overline{E}_{B_{1}}^q-\overline{E}_{B_{2}}^q)^2, \tag {6}
		\end{align}	
	\end{subequations}
	where, $\overline{E}_{B_{z}}^q = \frac{1}{v_z}\sum_{\hat{v}=1}^{v_z} E_{B_{z}}^{h, q}$ \cite{C_Hervada_Sala_program}, and $z = 1, 2$. It is noteworthy that the condition for merging two clusters into one cluster (denoted as $\widehat{B}$) is that the achieved $\bigtriangleup SSQ_{{B_{1}}{B_{2}}}$ for those two clusters is the smallest compared with all other cluster pairs. The process will repeat until reaching the stopping rule. 
    \par
    Regarding the \textit{stopping rule}, we consider it as the specified number of clusters, as HAC approach can be halted when the specified number of clusters is satisfied \cite{R_R_Yager_Intelligent_control}. \textit{However, since non-IID data setting is considered in this work, there may exist a phenomenon that the number of the embedding of each class of client $m$ (i.e., $|\mathcal{D}_{m, c}|$) is less than the specified number of clusters.} To cope with such an issue, CHAC is proposed by integrating the following condition into the original HAC method: \textit{the clustering method is performed only when $|\mathcal{D}_{m, c}|$ is greater than or equal to the specified number of clusters.} Let $\zeta_m^c$ denote the specified number of clusters, the corresponding additional constraint is given as below in this article:


	\begin{subequations}\label{Opt_7}
		\setlength{\abovedisplayskip}{3pt}
		\setlength{\belowdisplayskip}{3pt}
		\vspace{-0.45cm}
		\begin{align}
			\zeta_m^c = \left\{  
			\begin{array}{rcl}
				\widehat{\zeta}, & \textrm{if $|\mathcal{D}_{m, c}| \geq \widehat{\zeta}$}, \\
				|\mathcal{D}_{m, c}|,& \textrm{otherwise.}  \\
			\end{array} \right.  \tag{7}
		\end{align}
	\end{subequations} 
    Here $\widehat{\zeta}$ is a constant, while $\zeta_m^c=|\mathcal{D}_{m, c}|$ means each embedding will be considered as a cluster. 
    \par
    After determining the desired clusters, the centroid of each cluster is considered as each local prototype for that class \big(as shown in \eqref{Opt_3}\big). Namely, the number of local prototypes belonging to class $c$ of client $m$ is equal to the number of clusters of that class. Those local prototypes will be disseminated to DU, and then through DU, they will reach the central server for global prototype generation  (given in Section \ref{global_aggregation_objective}) 
	
    \vspace{-0.3cm}
	\subsection{A Novel Prototype Alignment (PA)}
    In this work, as outlined earlier, motivated by \cite{X_Yang_FedAS}, prototype alignment is designed to avoid useful information loss. Specifically, drawing on \cite{X_Yang_FedAS}, we let the global prototype after being received by each edge device to learn from the local embedding vectors obtained by using the local model gained in the previous round.
    
	
	\par
	Concretely, let $\mathcal{T} = \{1, 2, ..., t\}$ denote the set of rounds for training, for a given data $(x_m^\textrm{n}, y_m^\textrm{n}) \in \mathcal{D}_m$, we can give the local embedding of client $m$ by using the previous round local model as follows: 
	\begin{subequations}\label{Opt_8}
		\setlength{\abovedisplayskip}{3pt}
		\setlength{\belowdisplayskip}{3pt}
		\vspace{-0.1cm}
		\begin{align}
			\varrho_{m, c}^\textrm{n, t-1} = \Upsilon(x_m^\textrm{n};y_m^\textrm{n}=c, \theta_m^{t-1}), \tag{8}
		\end{align}
	\end{subequations} 
	where $t-1$ denotes the previous round, while $\theta_m^{t-1}$ indicates the parameter of the representation layer of client $m$ in the previous round. To align the global prototype with previous round local embedding vectors, for simplicity, we design the loss for prototype alignment (PA loss for short) based on MSE loss, which is given as below:
	\begin{subequations}\label{Opt_9}
		\setlength{\abovedisplayskip}{3pt}
		\setlength{\belowdisplayskip}{3pt}
		\begin{align}
			\mathcal{L}_m^\textrm{PA} = \frac{1}{|\mathcal{C}_m||\mathcal{D}_{m, c}|}\sum_{c=1}^{|\mathcal{C}_m|}\sum_{n=1}^{|\mathcal{D}_{m, c}|}(\varrho_{m, c}^\textrm{n, t-1}-\overline{P_{c}^t})^2, \tag{9}
		\end{align}
	\end{subequations} 
	where $\overline{P_{c}^t}$ is the global prototype at current round $t$, which can be computed by \eqref{Opt_14} (defined in Section \ref{global_aggregation_objective}). $|\mathcal{D}_{m, c}|$ denotes the number of the data in $\mathcal{D}_{m, c}$ that belongs to class $c$. $|\mathcal{C}_m|$ denotes the categories of client $m$.

   \vspace{-0.35cm}
   \subsection{LEMGP Loss Design}
   
	To form the local loss, in addition to the above losses (i.e., SKD loss, PA loss), we additionally introduce the cross-entropy (CE) loss and design a new loss function based on the COREL loss \cite{K_Kenyon_Clustering_Representation} (LEMGP loss as aforementioned). Specifically, CE loss, $\mathcal{L}_m^\textrm{CE}$, is the loss function that has widespread adoption in supervised learning \cite{P_Khosla_Supervised_Contrastive}. Regarding LEMGP loss (denoted as $\mathcal{L}_m^\textrm{LEMGP}$), it contains two parts: 1) attractive loss, $\mathcal{L}_m^\textrm{att}$, and 2) repulsive loss, $\mathcal{L}_m^\textrm{rep}$. In detail, we define it as:
	\begin{subequations}\label{Opt_10}
		\setlength{\abovedisplayskip}{3pt}
		\setlength{\belowdisplayskip}{3pt}
		\begin{align}
			\mathcal{L}_m^\textrm{LEMGP} =  \Lambda\mathcal{L}_m^\textrm{att} + (1-\Lambda)\mathcal{L}_m^\textrm{rep}, \tag{10}
		\end{align}
	\end{subequations} 
	where $\Lambda \in (0, 1]$ is a hyperparameter. 
	\par
	As for $\mathcal{L}_m^\textrm{att}$, it favors attracting the local embedding to the global prototype belonging to the same class. To do this, for simplicity, we design the attractive loss on the basis of MSE loss, and we call it as \textit{weighted MSE loss} in this work, which is given as below:
\begin{subequations}\label{Opt_11}
	\setlength{\abovedisplayskip}{3pt}
	\setlength{\belowdisplayskip}{3pt}
	\vspace{-0.1cm}
	\begin{align}
		\mathcal{L}_m^\textrm{att} = \sum_{c=1}^{|\mathcal{C}_m|}\lambda \frac{1}{|D_{m ,c}|}\sum_{n=1}^{|D_{m ,c}|} (\Upsilon(x_m^\textrm{n};y_m^\textrm{n}, \theta_m^{t})-\overline{P_{c}^t})^2, \tag{11}
	\end{align}
\end{subequations} 
where $\lambda$ is a hyperparameter. $\theta_m^{t}$ indicates the parameters of the representation layers at the current round $t$, while $\Upsilon(x_m^\textrm{n};y_m^\textrm{n}, \theta_m^{t})$ represents the local embedding obtained at the current round, when giving a data sample $(x_m^\textrm{n}, y_m^\textrm{n})$.
\par
In regard to the repulsive loss $\mathcal{L}_m^\textrm{rep}$, motivated by \cite{J_Deuschel_Multi_Prototype_Few_shot}, we define it as below:
\begin{subequations}\label{Opt_12}
	\setlength{\abovedisplayskip}{3pt}
	\setlength{\belowdisplayskip}{3pt}
	\begin{align}
		\mathcal{L}_m^\textrm{rep} = \log\sum_{c=1}^{|\mathcal{C}_m|}e^{-\lambda\frac{1}{|\mathcal{D}_m|}\sum_{n=1}^{|\mathcal{D}_m|}\big(\Upsilon(x_m^\textrm{n};y_m^\textrm{n}, \theta_m^{t})-\overline{P_{c}^t}\big)^2}. \tag{12}
	\end{align}
\end{subequations} 

\vspace{-0.2cm}
		
\subsection{Local Loss Function Design}
With regard to the local loss function, we devise it as a linear combination of CE loss, SKD loss, the proposed PA loss and the proposed LEMGP loss, which is defined by:
\begin{subequations}\label{Opt_13}
	\setlength{\abovedisplayskip}{3pt}
	\setlength{\belowdisplayskip}{3pt}
	\begin{align}
	\hspace{-0.3cm}	\mathcal{L}_m^\textrm{loss} \!=\! \mu_1\mathcal{L}_m^\textrm{CE} \!+\!  (1-\mu_1)\mathcal{L}_m^\textrm{SKD} \!+\! \mu_2\mathcal{L}_m^\textrm{PA} \!+\! \mu_3\mathcal{L}_m^\textrm{LEMGP}. \tag{13}
	\end{align}
\end{subequations} 
Here, $\mu_1$, $\mu_2$ and $\mu_3$ are three hyperparameters. Notably, in the $1^\textrm{st}$ round, we need to prepare a model to be the historical model of the next round of training. Thereby, only $\mathcal{L}_m^\textrm{CE}$ is employed in the $1^\textrm{st}$ round for simplicity. That is, $\mu_1=1$ and $\mu_2=\mu_3=0$.

\vspace{-0.2cm}

\subsection{Global Aggregation (Model \& Prototypes)}
\label{global_aggregation_objective}
\par
In this work, we consider both model aggregation and multi-prototype aggregation on the parameter server located on the CU layer in the considered AI-RAN enabled MEC system to generate the global model and the global prototypes, respectively. Regarding the global model generation, \eqref{Opt_2} is employed. As for the global prototype of class $c$ in each round, as per \eqref{Opt_4}, it can be given as follows: 
	\begin{subequations}\label{Opt_14}
	\setlength{\abovedisplayskip}{3pt}
	\setlength{\belowdisplayskip}{3pt}
	\begin{align}
		\overline{P_{c}} = \frac{1}{|\mathcal{M}_c|\zeta_m^c}\sum_{m=1}^{\left |\mathcal{M}_c\right |}\sum_{i=1}^{\zeta_m^c} \frac{\left| \mathcal{D}_{m, c} \right|}{\left| \mathcal{D}_c \right|}P_{m, c}^{i}.  \tag{14} 
	\end{align}	
\end{subequations}

\vspace{-0.2cm}
 	
\subsection{Algorithm Summary}
In this part, we summarize the proposed approach in Algorithms \ref{alg_1} and \ref{Opt_2}, respectively. Particularly, Algorithm \ref{alg_1} exhibits the multi-prototype generation for each class on the local edge device via the presented CHAC approach. Algorithm \ref{alg_2} clarifies the process of the proposed MP-FedKD. Next, the details will be illustrated.

 \begin{figure}[t!]
	\vspace{-0.3cm}	
	\begin{algorithm}[H]	
		\renewcommand{\algorithmicrequire}{\textbf{Input:}}
		\renewcommand{\algorithmicensure}{\textbf{Output:}}
		\caption{Multi-prototype generation via the conditional HAC (each client)}
		\label{alg_1}
		\begin{algorithmic}[1]
			\REQUIRE Local embedding of client $m$ belonging to class $c$: $\varrho_{m, c}^\textrm{1}$, $\varrho_{m, c}^\textrm{2}$, ..., $\varrho_{m, c}^\textrm{n}$
			\ENSURE  Composition of each cluster
			\STATE Set stopping rule: \eqref{Opt_7}
                \STATE Assign each local embedding of class $c$, $\varrho_{m, c}^\textrm{n}$, to each single cluster
                \IF{$|\mathcal{D}_{m, c}| \geq \widehat{\zeta}$}  	
			\REPEAT
			\STATE Compute $\bigtriangleup SSQ_{{B_{1}}{B_{2}}}$ using \eqref{Opt_6}
			\STATE Merge two clusters with the smallest $\bigtriangleup SSQ_{{B_{1}}{B_{2}}}$
			\STATE Obtain new clusters
			\UNTIL{Get $\zeta_m^c=\widehat{\zeta}$ clusters}
            \ELSE 
            \STATE Get $\zeta_m^c=|\mathcal{D}_{m, c}|$ clusters
                \ENDIF
			\STATE \textbf{return} Composition of $\zeta_m^c$ cluster
		\end{algorithmic}  
	\end{algorithm} 
	\vspace{-0.85cm}
\end{figure}	

\par
In Algorithm \ref{alg_1}, in line $1$, we use \eqref{Opt_7} to set the stopping rule to end the execution of CHAC method on the local client. In line $2$, we allocate each local embedding per class $c$, $\varrho_{m, c}^\textrm{n}$, to each single cluster. Then, we check whether the number of data is greater or equal to the specified cluster number. If the condition is satisfied (line $3$), we will repeat lines $5-7$ until reaching the stopping rule (i.e., obtain $\zeta_m^c$ clusters) in line $8$. To be specific, in line $5$, we adopt \eqref{Opt_6} to compute $\bigtriangleup SSQ_{{B_{1}}{B_{2}}}$, and then two clusters with the smallest $\bigtriangleup SSQ_{{B_{1}}{B_{2}}}$ are merged in line $6$ so as to get new clusters in line $7$. If the condition is not met, then each local embedding of class $c$ will be regarded as each independent cluster, viz., $\zeta_m^c=|\mathcal{D}_{m, c}|$ clusters will be obtained (line $10$). In line $12$, we get the composition of each cluster. 

\begin{figure}[t!]
	\vspace{-0.25cm}
	\begin{algorithm}[H]	
		\renewcommand{\algorithmicrequire}{\textbf{Input:}}
		\renewcommand{\algorithmicensure}{\textbf{Output:}}
		\caption{Proposed MP-FedKD Approach}
		\label{alg_2}
			\begin{algorithmic}[1]
				\REQUIRE $\mathcal{D}_m=\{(x_m^\textrm{1}, y_m^\textrm{1}), (x_m^\textrm{2}, y_m^\textrm{2}), ..., (x_m^\textrm{n}, y_m^\textrm{n})\}$,
				\ENSURE Global Model $\omega$ \\
				\hspace{-0.55cm} \textbf{Process on the central server (CU layer):}
				\STATE Initialize global model (i.e., $\omega$) and global prototype set (i.e., $\mathbb{P} = \{\overline{P_{1}}, \overline{P_{2}}, ..., \overline{P_{c}}\}$)	
				\FOR{each round $t = 1, 2, ..., T$}
				\FOR{each participating client}
				\IF {$t == 1$} 	    
				\STATE $\omega_m^t, \mathbb{P}_m \leftarrow$ DU $\leftarrow \textrm{ClientUpdate}(\omega, \mathbb{P}, t=1)$
				\ELSE 
				\STATE $\omega_m^t, \mathbb{P}_m \leftarrow$ DU $\leftarrow \textrm{ClientUpdate}(\omega, \mathbb{P}, t)$
				\ENDIF
				\ENDFOR
				\STATE Obtain global prototype set $\mathbb{P}$ via  \eqref{Opt_14}
				\STATE Obtain global model $\omega$ via  \eqref{Opt_2} 
				\ENDFOR
				\STATE \textbf{return} $\omega$
				\hfill \break
				\\
				\hspace{-0.63cm} \textbf{ClientUpdate}($\omega$, $\mathbb{P}, t$):
				\STATE Replace local model with the global model: $\omega_m \leftarrow \omega$
				\FOR{each epoch}
				\FOR{batch $(x_m, y_m) \in \mathcal{D}_m$}
				\STATE Obtain multi-prototype per class via Algorithm \ref{alg_1}
				\STATE Compute CE loss $\mathcal{L}_m^\textrm{CE}$
				\IF {$t \neq 1$}
					\STATE Compute SKD loss $\mathcal{L}_m^\textrm{SKD}$ via  \eqref{Opt_5}
					\STATE Compute PA loss $\mathcal{L}_m^\textrm{PA}$ via  \eqref{Opt_9}
					\STATE Compute LEMGP loss $\mathcal{L}_m^\textrm{LEMGP}$ via  \eqref{Opt_10}
				\ENDIF			
				\STATE Compute local loss $\mathcal{L}_m^\textrm{loss}$ via \eqref{Opt_13}
				\STATE $\omega_m \leftarrow \omega_m - \delta\bigtriangledown\mathcal{L}_m^\textrm{loss}$ \cite{H_B_McMahan_Communication_Decentralized}
				\ENDFOR
				\ENDFOR	
				\STATE Gain the updated local model ($\omega_m^t$) and a set of local prototypes ($\mathbb{P}_m = \{\{P_{m, c}^i\}_{i=1}^{|\mathcal{P}_{m, c}|}, c \in \mathcal{C}_m\}$) at round $t$
				\STATE \textbf{return} $\omega_m^t$, $\mathbb{P}_m$
			\end{algorithmic}  
		\end{algorithm} 
		\vspace{-0.95cm}
	\end{figure}

\par 
In Algorithm \ref{alg_2}, in lines $1-13$, the process executed on the central server (placed on the CU layer) is provided. Specifically, in line $1$, both the global model (i.e., $\omega$) and global prototypes (i.e., $\mathbb{P} = \{\overline{P_{1}}, \overline{P_{2}}, ..., \overline{P_{c}}\}$) will be initialized. Then in each round (line $2$), the central server will receive both the updated local model and the local prototypes through the relay of DU (line $5$ or line $7$) from each participating client (line $3$). Notably, line $5$ represents the process of the first round ($t=1$), while line $7$ shows the process of other rounds ($t\neq1$). It is worth mentioning that in both line $5$ and line $7$, the client will be updated based on the received global model $\omega$ and the global prototype set $\mathbb{P}$, which are relayed by DU. In line $10$, we use \eqref{Opt_14} to get the global prototype set, $\mathbb{P}$. In line $11$, the global model, $\omega$, is calculated via \eqref{Opt_2}. The local client update process is shown in lines $14-29$. In particular, in line $14$, we replace the local model (i.e., $\omega_m$) with the global model (i.e., $\omega$). For each batch (line $16$) per epoch (line $15$), we utilize Algorithm \ref{alg_1} to get multiple local prototypes for each class in line $17$. In line $18$, CE loss will be computed. If it is not the first round, i.e., $t\neq 1$ (line $19$), we will calculate SKD loss (line $20$), PA loss (line $21$) and LEMGP loss (line $22$). The local loss will be obtained in line $24$. In line $25$, the local model will be updated via $\omega_m \leftarrow \omega_m - \delta\bigtriangledown\mathcal{L}_m^\textrm{loss}$ \cite{H_B_McMahan_Communication_Decentralized}, where $\delta$ represents the learning rate. In line $28$, we will get the updated local model (denoted as $\omega_m^t$) and the updated local prototypes per class at current round $t$, which will be submitted to DU and then relayed to the central server (in CU layer).
	
\par
\textit{\textbf{Time Complexity}}: The time complexity of the proposed MP-FedKD approach is analyzed based on the following main aspects: 1) global model aggregation (GMoA) on the central server, 2) global prototype aggregation (GProA) on the central server, 3) multi-prototype generation on the local client side (MPG-LC), and 4) the backbone network. Regarding GMoA, as aforementioned, FedAvg is leveraged to produce the global model. Among a set of clients, $\mathcal{M}$, consider a total of $M'$ participating clients join the training process, where $M' \leq |\mathcal{M}|$. The corresponding time complexity for performing FedAvg in the central server side is bounded as $O(M')$ \cite{X_Li_FedGTA}. In the case of GProA, in each round, if single-prototype strategy is adopted, the corresponding time complexity of GProA does not exceed $O(|\mathcal{C}|M')$ \cite{L_Zou_Cyber_Attacks_Prevention_Prosumer_EV}. Accordingly, we can infer that, for this work (multi-prototype strategy is considered), the time complexity of GProA is not greater than $O(|\mathcal{C}|\sum_{m=1}^{M'}\zeta_m^c)$. 
As a result, the time complexity for the \textit{\textbf{central server}} side is no more than $O(M'+|\mathcal{C}|\sum_{m=1}^{M'}\zeta_m^c)$.
Concerning MPG-LC, the designed CHAC is performed on each client to generate the local prototypes of each class. Since CHAC approach is constructed based on HAC approach, the time complexity for client $m$ to produce the local prototypes belonging to class $c$ can be given as $O(|\mathcal{D}_{m, c}|^3)$ as per \cite{S_Sieranoja_Fast_agglomerative_clustering}. Accordingly, for all the categories of local client (i.e., $\mathcal{C}_m=\{1, 2, ..., C_m\}$), the time complexity for generating all local prototypes on client $m$ can be obtained by  $O(\sum_{c=1}^{|\mathcal{C}_m|}|\mathcal{D}_{m, c}|^3)$. As for the backbone network's time complexity, it will depend on the backbone network being used. For instance, consider the combination of the convolutional neural network (CNN) and the fully connected network (FCN) as an example to analyze the time complexity. The time complexity of CNN is $O(\sum_{j} e_{j-1} \cdot s_j^2 \cdot e_j \cdot a_j^2)$ \cite{K_He_CNN_Constrained_Time}. Here, $j$ denotes the convolutional layer's index. $e_{j-1}$ represents the number of the input channels for the $j^\textrm{th}$ layer, and $e_j$ denotes the filters' number in the $j^\textrm{th}$ layer. $s_j$ indicates the filter's spatial size, and $a_j$ denotes the output feature's spatial size. With respect to FCN, the time complexity is $O(\sum_{z}\hat{e}_{z-1}\hat{e}_{z})$, where $\hat{e}_{z-1}$ represents the number of neural units in the ${z-1}^\textrm{th}$ layer of FCN, while $\hat{e}_{z}$ denotes the neural units' number in the $z^\textrm{th}$ layer of FCN. Thus, motivated by \cite{L_Zou_Cyber_Attacks_Prevention_Prosumer_EV}, the time complexity for each local client can be given as $O\Big(\widetilde{E}\big(\sum_{c=1}^{|\mathcal{C}_m|}|\mathcal{D}_{m, c}|^3+\sum_{j} e_{j-1} \cdot s_j^2 \cdot e_j \cdot a_j^2+\sum_{z}\hat{e}_{z-1}\hat{e}_{z}\big)\Big)$, where $\widetilde{E}$ indicates the local iteration. Accordingly, inspired by \cite{C_Pan_Fair_Graph_FL}, the total time complexity of the proposed MP-FedKD method for each \textit{\textbf{local client}} can be considered to be at most $O\Big(\widetilde{E}\big(\sum_{c=1}^{|\mathcal{C}_m|}|\mathcal{D}_{m, c}|^3+\sum_{j} e_{j-1} \cdot s_j^2 \cdot e_j \cdot a_j^2+\sum_{z}\hat{e}_{z-1}\hat{e}_{z}\big) + M' + |\mathcal{C}|\sum_{m=1}^{M'}\zeta_m^c\Big)$.

\section{Experimental Analysis}
\par
In this section, we delve into the experimental analysis for the proposed MP-FedKD towards the considered AI-RAN enabled MEC system in several aspects, such as backbone network and clustering evaluation, scalability analysis, robustness evaluation, and ablation study. Next, details will be offered.

\subsection{Experimental Setup}
\par
\textbf{\textit{Dataset: }} In the experiment, we employ six datasets with Non-IID settings to evaluate the performance of the proposed MP-FedKD for the considered AI-RAN enabled MEC system. Those datasets are CIFAR-10 \cite{A_Krizhevsky_Learning_multiple_tiny}, MNIST, Fashion-MNIST \cite{H_Xiao_Fashion_MNIST}, EuroSAT \cite{P_Helber_EuroSAT}, \textit{\textbf{M+F}}, and \textit{\textbf{C+E}}, respectively. Concretely, as per \cite{Demystifying_Impact_Key_FL}, \textit{\textbf{CIFAR-10}} owns $60, 000$ $32\times32$ color images from $10$ categories, where $50, 000$ images are used for training and $10, 000$ images are utilized for test. As for \textit{\textbf{MNIST}} and \textit{\textbf{Fashion-MNIST}}, both of them contain images with image size $28\times28$ belonging to $10$ categories ($60, 000$ images for training, while $10, 000$ images for test). Regarding \textit{\textbf{EuroSAT}}, it is a dataset for the purpose of land use and land cover classification, which is composed of $27, 000$ labeled images across 10 classes \cite{P_Helber_EuroSAT}. Each image of EuroSAT is $64\times64$ pixels. For the EuroSAT dataset, we randomly choose $70\%$ of the images as the training set and regard the remaining $30\%$ as the test set. In the case of \textit{\textbf{M+F}} dataset, we combine MNIST and Fashion-MNIST as one dataset. In regard to \textit{\textbf{C+E}} dataset, it is composed of CIFAR-10 dataset and EuroSAT dataset. It is worth mentioning that it is feasible to make such a combination, as the images of both MNIST and Fashion-MNIST are $1$ channel images, and the images of CIFAR-10 and EuroSAT have $3$ channels.

\par
\textbf{\textit{Non-IID Settings:}} For evaluation, we take into account two manners to prepare the non-IID data, as shown below:
\begin{itemize}
	\item \textbf{Non-IID type 1 (Same Domain)}: In this manner, each client (i.e., edge device)'s local dataset comes from the same domain (i.e., CIFAR-10 \cite{A_Krizhevsky_Learning_multiple_tiny}, MNIST, Fashion-MNIST \cite{H_Xiao_Fashion_MNIST} or EuroSAT \cite{P_Helber_EuroSAT}). In addition, Dirichlet Distribution with several parameters are employed to prepare the non-IID data. Those parameters are $Dir = \{0.3, 0.5, 0.7, 0.9\}$.
	\item \textbf{Non-IID type 2 (Distinct Domain)}: In this case, \textit{M+F} dataset and \textit{C+E} dataset are leveraged. To make the non-IID data for \textit{C+E} dataset, firstly, we randomly divide all clients into two groups. Secondly, we let one group's data come from one domain, and let the other group's data come from the other domain. Dirichlet Distribution ($Dir = \{0.3, 0.5, 0.7, 0.9\}$) is performed on each group. The same process is also adopted for \textit{M+F} dataset.
\end{itemize}
\par
\textbf{\textit{Model Setup:}}
We consider three models, i.e., simple CNN (S-CNN), ResNet-8 and ResNet-10, to form each local client. For S-CNN, we consider to construct it by two convolutional layers, and three fully connected (FC) layers. The input channel for MNIST and Fashion-MNIST is $1$, and for CIFAR10 and EuroSAT is $3$. The output channel of the first convolutional layer is $64$, which is the second convolutional layer's input channel. The output channel of the second convolutional layer is $128$. As for the three FC layers, the (input channel, output channel) pair for each layer is given as ($128\times \textrm{image size}$, $64$), ($64$, $32$), and ($32$, $10$). As for ResNet-8 and ResNet-10, identical to \cite{Z_Liu_Privacy_Split_Learning_Ensembles}, ResNet-8 is composed of a single convolutional layer, 3 basic blocks (two convolutional layers per block), and 1 FC layer. With respect to ResNet-10, compared to ResNet-8, it contains an additional block (two convolutional layers) before the FC layer. Other parameters are defined as: batch size is set as $32$, learning rate is $0.001$, local epoch is $5$, overall communication rounds are set as $50$, temperature $\tau$ is given as $0.1$ \cite{T_Liang_Compressing_Multiobject}, hyperparameters $\mu_1$, $\mu_2$, and $\mu_3$ are separately set as $0.9$, $1$ and $0.1$, $\Lambda=\lambda=0.5$ \cite{J_Deuschel_Multi_Prototype_Few_shot}.

\par
\textbf{\textit{Baseline:}}
For comparison, we take into consideration the following FL-based baselines: FedProx \cite{T_Li_FedProx}, FedProto \cite{Y_Tan_FedProto}, FedAS \cite{X_Yang_FedAS}, MOON \cite{Q_Li_MOON}, E-FPKD \cite{L_Zou_Cyber_Attacks_Prevention_Prosumer_EV}, and FedALA \cite{J_Zhang_FedALA}. To benchmark the CHAC approach, we consider a baseline where we replace the CHAC of the proposed MP-FedKD with K-Means clustering. Note that we consider a random client selection mechanism in each round for all methods for simplicity. To ensure fairness, all baselines follow the model architecture of the proposed MP-FedKD approach. 

\vspace{-0.2cm}
\subsection{Backbone Network and Clustering Evaluation}
\par
In Table \textcolor{blue}{\ref{various_backbone_ACC_MAE_MFS}}, we demonstrate the accuracy (ACC), mean absolute error (MAE) and Macro F1 Score (MFS) achieved by the proposed MP-FedKD approach, where three backbone networks (i.e., S-CNN, ResNet-8 and ResNet-10) and two datasets (i.e., CIFAR10 and Fashion-MNIST) with $Dir=0.9$ are taken into account. It can be observed that the proposed method with ResNet-10 as backbone can achieve the highest ACC, highest MFS and the lowest MAE. Thus, since ResNet-10 can achieve the best values, for convenience, we choose ResNet-10 to form each local client for further evaluation.

\begin{table}[t!]
	\caption{Various backbone comparisons by considering accuracy (ACC), MAE, and Macro F1 Score (MFS).}
    \vspace{-0.2cm}
	\renewcommand\arraystretch{1}
	\begin{center}
		\begin{tabular}{|c|c|m{1.1cm}|m{1.1cm}|m{1.1cm}|}
			\hline
			\hfil \textbf{Dataset} & \hfil \textbf{Backbone} & \hfil \textbf{ACC} & \hfil \textbf{MAE} &  \hfil  \textbf{MFS}\\ \hline
			\hfil \multirowcell{3}{CIFAR10}& \hfil S-CNN & \hfil $0.4948$ & \hfil $1.8022$ & \hfil $0.4944$ \\ \cline{2-5}
			& \hfil ResNet-8 & \hfil $0.6769$ & \hfil $1.1301$ & \hfil $0.6758$ \\ \cline{2-5}	
			& \hfil ResNet-10 & \hfil $\textbf{0.6862}$  & \hfil $\textbf{1.1033}$ & \hfil $\textbf{0.6823}$ \\ 		
			\hline \hline	
			\multirowcell{3}{Fashion-MNIST}& \hfil S-CNN & \hfil $0.8801$ & \hfil $0.3865$ & \hfil $0.8781$ \\ \cline{2-5}	
			& \hfil ResNet-8 & \hfil $0.9007$  & \hfil $0.3281$ & \hfil $0.9015$ \\ \cline{2-5}	
			& \hfil ResNet-10 & \hfil $\textbf{0.9097}$  & \hfil $\textbf{0.2982}$ & \hfil $\textbf{0.9088}$ \\  \hline							
		\end{tabular}
		\label{various_backbone_ACC_MAE_MFS}
	\end{center}
    \vspace{-0.4cm}
\end{table}

\par
In Table \textcolor{blue}{\ref{table_ours_various_cluster_number}}, we evaluate the effect of different cluster numbers, $\zeta_m^c$, on the performance of the proposed method, where $Dir=0.9$. Compared to $\zeta_m^c=2$ and $\zeta_m^c=4$, when $\zeta_m^c=3$, the ACC and root mean square error (RMSE) achieved on most datasets are the best. Accordingly, for simplicity, we adopt $\zeta_m^c=3$ for the proposed method in the following evaluation.

\begin{table}[t!]
	\caption{Accuracy and RMSE achieved by the proposed method under various cluster numbers (i.e., $\zeta_m^c$).}
	\renewcommand\arraystretch{1}
	\centering
	\scalebox{0.9}{
	\begin{tabular}{cccccccc}
		\toprule
		\multirow{2}{*}{\textbf{Dataset}} & \multicolumn{2}{c}{$\zeta_m^c=2$} & \multicolumn{2}{c}{$\zeta_m^c=3$} & \multicolumn{2}{c}{$\zeta_m^c=4$} \\
		\cmidrule(rl){2-3} \cmidrule(rl){4-5} \cmidrule(rl){6-7}
		 & {ACC} & {RMSE} & {ACC} & {RMSE} & {ACC} & {RMSE} \\
		\midrule
		CIFAR-10 & $\textbf{0.6776}$ & $\textbf{2.4146}$ & $0.6710$ & $2.4347$ & $0.6390$ & $2.5070$ \\
		MNIST & $0.9914$ & $0.3872$ & $\textbf{0.9933}$ & $\textbf{0.3415}$ & $0.9929$ & $0.3833$ \\
		Fashion-MNIST & $0.9075$ & $1.1463$ & $\textbf{0.9097}$ & $\textbf{1.1277}$ & $0.9059$ & $1.1846$ \\
		EuroSAT & $0.8413$ & $1.6649$ & $0.8390$ & $1.7148$ & $\textbf{0.8556}$ & $\textbf{1.6456}$ \\
		M+F & $0.9350$ & $0.9706$ & $\textbf{0.9392}$ & $\textbf{0.9218}$ & $0.9288$ & $1.0146$ \\
		C+E & $0.5356$ & $3.4547$ & $\textbf{0.5897}$ & $\textbf{2.9880}$ & $0.5702$ & $3.2205$ \\
		\bottomrule
	\end{tabular}
	\label{table_ours_various_cluster_number}
}
\end{table}

\begin{figure} [t!]
	\centering
	\includegraphics[scale = .42]{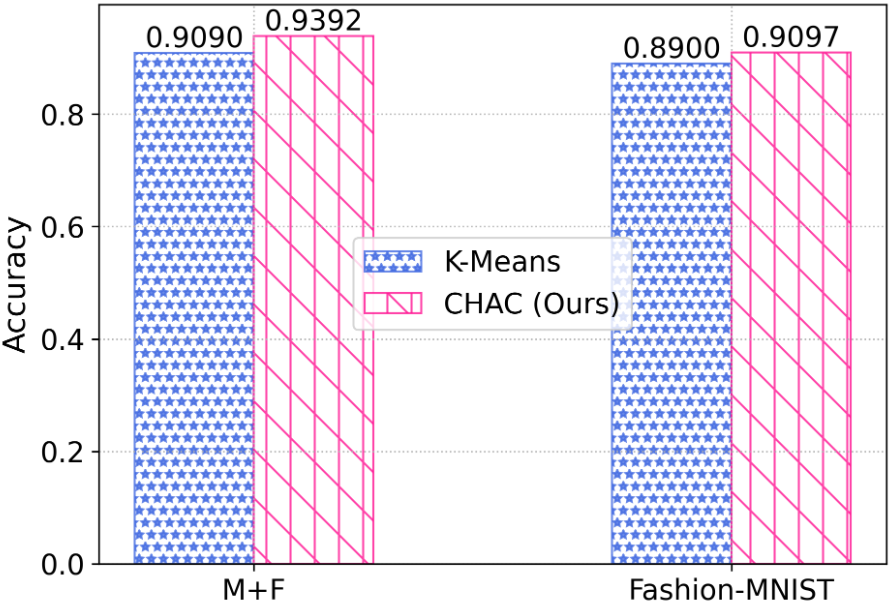}
	\captionsetup{font=small}
	\caption{Accuracy comparison between K-Means and CHAC (Ours).}
	\label{fig_K_Means_Ours_Clustering}
	\vspace{-0.4cm}
\end{figure}

\par
In Fig. \textcolor{red}{\ref{fig_K_Means_Ours_Clustering}}, to evaluate the performance of the CHAC approach, we consider comparing it with K-Means-based MP-FedKD approach, where M+F dataset and fashion-MNIST dataset are leveraged. From Fig. \textcolor{red}{\ref{fig_K_Means_Ours_Clustering}}, it can be seen that the accuracy achieved by the CHAC-based MP-FedKD approach is approximately $1.03\times$ and $1.02\times$ greater than K-Means-based MP-FedKD on the M+F dataset and Fashion-MNIST dataset, respectively. The reason for this phenomenon is speculated to be: the CHAC is presented based on HAC approach, which is a hierarchical approach. As per \cite{H_R_A_Putri_segmentation_hierarchical_clustering}, the hierarchical approach can offer more information than the non-hierarchical method (e.g., K-Means \cite{J_Park_Development_WEEE}), owing to the employed dendrogram.

\vspace{-0.2cm}

\subsection{Scalability and Error Analysis}
\begin{figure*}[t!]
	\centering
	\begin{subfigure}{0.245\textwidth} 
		\includegraphics[width=1 \linewidth]{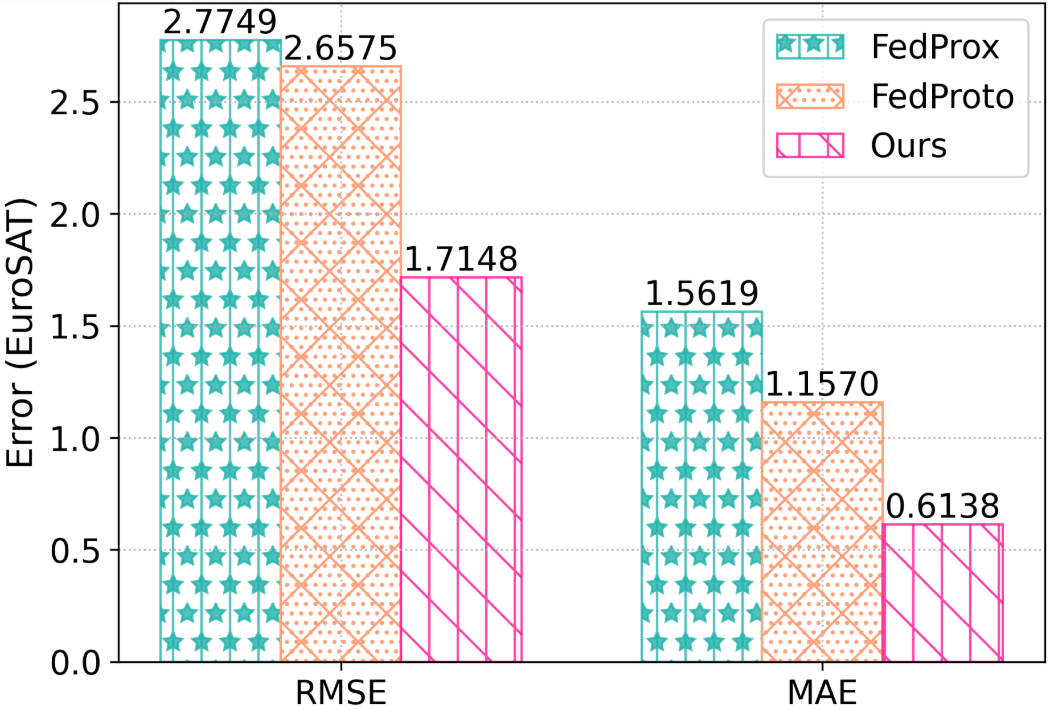}
		\captionsetup{font=small}
		\caption{EuroSAT}
		\label{fig_error_EuroSAT}
	\end{subfigure} 
	\begin{subfigure}{0.245\textwidth} 
		\includegraphics[width=1 \linewidth]{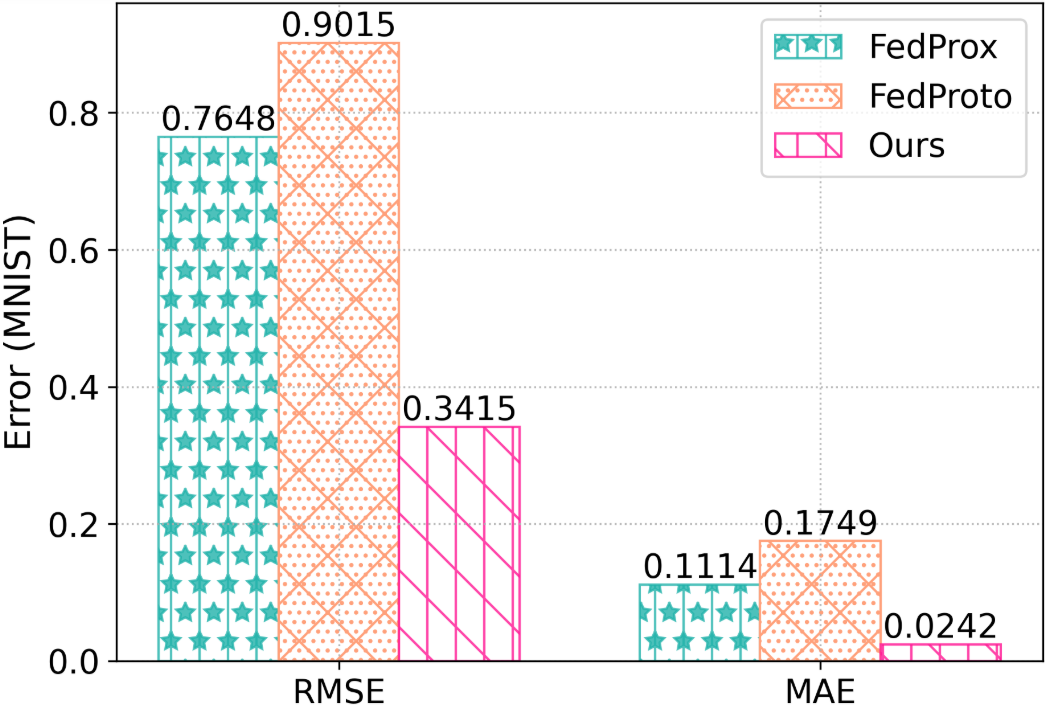}
		\captionsetup{font=small}
		\caption{MNIST}
		\label{fig_error_MNIST}
	\end{subfigure}
	\begin{subfigure}{0.245\textwidth} 
		\includegraphics[width=1 \linewidth]{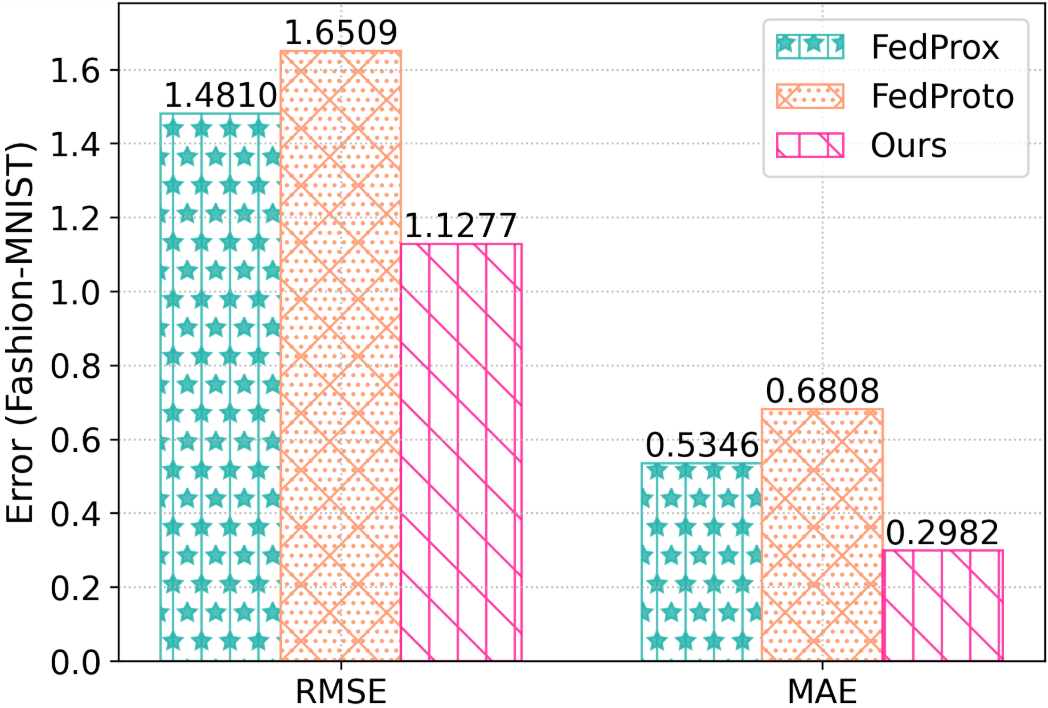}
		\captionsetup{font=small}
		\caption{Fashion-MNIST}
		\label{fig_error_Fashion_MNIST}
	\end{subfigure} 
	\begin{subfigure}{0.245\textwidth} 
		\includegraphics[width=1 \linewidth]{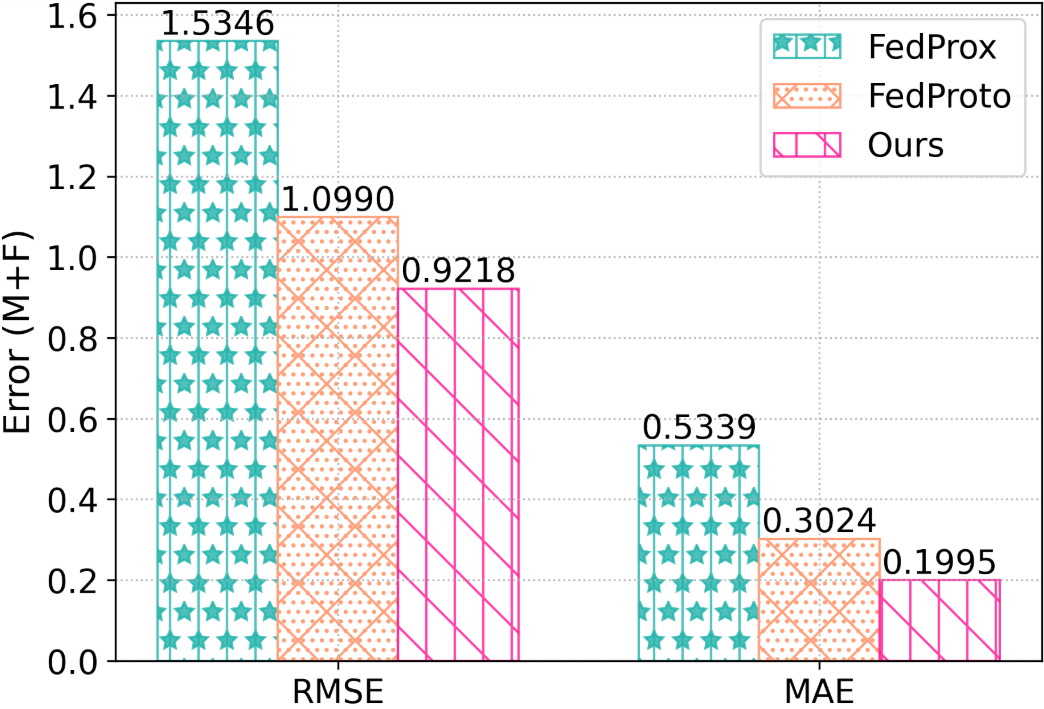}
		\captionsetup{font=small}
		\caption{M+F}
		\label{fig_error_M_F}
	\end{subfigure} 
		\vspace{-0.5cm}
	\captionsetup{font=small}
	\caption {RMSE and MAE achieved by FedProx, FedProto and the proposed method using various datasets ($Dir=0.9$).}
	\label{fig_error_Various_Datasets}
\end{figure*}

\par
In Fig. \textcolor{red}{\ref{fig_error_Various_Datasets}}, in order to reveal the superiority of the proposed MP-FedKD approach, We conduct a comparison experiment among FedProx, FedProto, and the proposed method. As we can see from Fig. \textcolor{red}{\ref{fig_error_Various_Datasets}}, the proposed method can realize the smallest RMSE and MAE on the considered four datasets. Specifically, in Fig. \textcolor{red}{\ref{fig_error_EuroSAT}}, for EuroSAT dataset, the RMSE achieved by the proposed method is roughly $1.62\times$ and $1.55\times$ less than FedProx and FedProto, respectively, while the MAE realized by the proposed method is separately around $2.54\times$ and $1.88\times$ smaller than FedProx and FedProto. In Fig. \textcolor{red}{\ref{fig_error_MNIST}}, Fig. \textcolor{red}{\ref{fig_error_Fashion_MNIST}} and Fig. \textcolor{red}{\ref{fig_error_M_F}}, it can also be found that the RMSE and MAE obtained by the proposed method are the lowest. Consequently, it can be concluded that the proposed method outperforms the other two methods. The reasons for this phenomenon are conjectured as follows: 1) in comparison with FedProx and FedProto, SKD is adopted to be a part of the solution. According to \cite{J_Tang_FedRAD}, it is effective to address data heterogeneity when integrating KD into FL. As the considered SKD belongs to KD, we can infer that using SKD can help boost the performance of the proposed method. 2) Instead of the single prototype-based strategy, the multi-prototype-based strategy is adopted in this work to build the solution, which can help the model grasp the samples' feature information comprehensively. 3) Owing to prototype alignment scheme in MP-FedKD approach, the global prototypes are allowed to learn from the previous round of local embedding vectors. Owing to this, prototype alignment process may help the proposed MP-FedKD approach work more effectively, thereby improving the performance.

\par
To further exhibit the effectiveness of the proposed MP-FedKD approach, scalability comparison experiments are conducted, as shown in Tables \textcolor{blue}{\ref{Scalability_single_dataset}} and \textcolor{blue}{\ref{Scalability_heterogeneous_datasets}}. Particularly, homogeneous datasets are considered in Table \textcolor{blue}{\ref{Scalability_single_dataset}}, while heterogeneous datasets (i.e., M+F dataset and C+E dataset) are employed in Table \textcolor{blue}{\ref{Scalability_heterogeneous_datasets}}. In addition, identical to \cite{M_Abdel_Basset_Lightweight_Convolutional}, which assesses the scalability of its proposed approach by taking into account different clients' numbers, in this article, we consider conducting the proposed method on different scales of clients to evaluate its scalability. Particularly, we consider $3$ cases which involve $10$, $20$ and $50$ clients, respectively, and $Dir=0.9$ is taken into consideration. Through those two Tables, it can be known that the accuracy realized by the proposed method is the highest in all cases regardless of the number of the clients, compared with the baselines. To be specific, taking the MNIST dataset and the number of clients as $10$, we can obtain that the accuracy achieved by the proposed method outperforms FedProx, FedAS, E-FPKD, MOON and FedProto by $2.13\%$, $0.87\%$, $6.15\%$, $0.31\%$, $4.06\%$, respectively. Notably, we notice that in most cases of Tables \textcolor{blue}{\ref{Scalability_single_dataset}} and \textcolor{blue}{\ref{Scalability_heterogeneous_datasets}}, the accuracy is the largest when the number of clients is $10$, compared with other cases. The reasons are speculated as: 1) the overall data amount of those datasets is fixed despite the number of clients; 2) for each round, a random client selection scheme is introduced, thereby there may be not all clients join the training process per round. However, since the proposed method is compatible with a different number of clients, it is still reasonable to infer that the proposed method has the scalability ability. For convenience, $10$ clients will be considered in the following assessment.

\begin{table}[t!]
	\renewcommand{\thetable}{\Roman{table}}
	\renewcommand\arraystretch{1}
	\setlength{\extrarowheight}{1 pt}
	\caption{Accuracy comparison with considering various number of clients using CIFAR10, MNIST, Fashion-MNIST and EuroSAT datasets.}
	\begin{center}
		\begin{tabular}{c | c |>{\centering} *3{m{1.3cm}}}
			\Xhline{0.8pt} 
			\toprule
			\multirow{2.1}{*}{\textbf{Dataset}}& \multirow{2.1}{*}{\textbf{Method}} &\multicolumn{3}{c}{\textbf{Accuracy}} \\\cline{3-5}
			& &  \multirow{1.5}{*}{\bfseries \# 10} & \multirow{1.5}{*}{\bfseries \# 20} & \multirow{1.5}{*}{\bfseries \# 50} \\
			\midrule
			\multirow{6}{*}{CIFAR10} & FedProx & \hfil $0.3733$ & \hfil $0.3155$ & \hfil $0.3223$ \\ 
			& FedAS & \hfil $0.4816$ & \hfil $0.4234$ & \hfil $0.4037$ \\
			& E-FPKD & \hfil $0.4494$ & \hfil $0.3501$ & \hfil $0.3146$ \\
			& MOON & \hfil $0.6183$ & \hfil $0.4302$ & \hfil $0.3700$ \\  
			& FedProto  & \hfil $0.4458$  & \hfil $0.3853$ & \hfil $0.4109$ \\	
			& \mycc Ours & \hfil \mycc $\textbf{0.6710}$ & \hfil \mycc $\textbf{0.6355}$ & \hfil \mycc $\textbf{0.5701}$ \\  
			\midrule
			\multirow{6}{*}{MNIST} & FedProx & \hfil $0.9720$ & \hfil $0.9240$ & \hfil $0.8530$ \\
			& FedAS & \hfil $0.9846$ & \hfil $0.9759$  & \hfil $0.9631$ \\
			& E-FPKD & \hfil $0.9318$  & \hfil $0.9309$ & \hfil $0.8849$ \\
			& MOON & \hfil $0.9902$  & \hfil $0.9679$ & \hfil $0.9442$ \\   
			& FedProto & \hfil $0.9527$ & \hfil $0.8998$ & \hfil $0.8627$ \\
			& \mycc Ours & \hfil \mycc $\textbf{0.9933}$ & \hfil \mycc $\textbf{0.9911}$ & \hfil \mycc $\textbf{0.9900}$ \\  
			\midrule
			\multirow{6}{*}{Fashion-MNIST} & FedProx &  \hfil $0.8317$ & \hfil  $0.7588$ & \hfil $0.6983$\\
			& FedAS & \hfil $0.8272$  & \hfil $0.8310$ & \hfil $0.8210$ \\
			& E-FPKD & \hfil $0.8087$  & \hfil $0.7689$ & \hfil $0.7203$ \\
			& MOON & \hfil $0.8892$ & \hfil $0.8338$ & \hfil $0.6852$ \\   
			& FedProto & \hfil $0.7826$ & \hfil $0.7453$ & \hfil $0.7444$ \\
			& \mycc Ours & \hfil \mycc $\textbf{0.9097}$ & \hfil \mycc $\textbf{0.8965}$ & \hfil \mycc $\textbf{0.8245}$ \\  
			\midrule
			\multirow{6}{*}{EuroSAT} & FedProx & \hfil $0.5910$ & \hfil $0.5513$ & \hfil $0.5156$ \\
			& FedAS & \hfil $0.7635$ & \hfil $0.6786$ & \hfil $0.6839$ \\
			& E-FPKD & \hfil $0.5520$  & \hfil $0.4578$  & \hfil $0.5311$ \\
			& MOON & \hfil $0.8192$ & \hfil $0.7401$ & \hfil $0.6616$ \\   
			& FedProto & \hfil $0.7553$ & \hfil $0.7423$ & \hfil $0.7498$ \\
			& \mycc Ours & \hfil \mycc $\textbf{0.8390}$ & \hfil \mycc $\textbf{0.8346}$ & \hfil \mycc $\textbf{0.8925}$ \\  
			\bottomrule
			\hline
		\end{tabular}
	\end{center}
	\label{Scalability_single_dataset}
		\vspace{-0.5cm}
\end{table}

\vspace{-0.2cm}

\subsection{Robustness Evaluation}
\par
\begin{figure*}[t!]
	\centering
	\begin{subfigure}{0.245\textwidth} 
		\includegraphics[width=1 \linewidth]{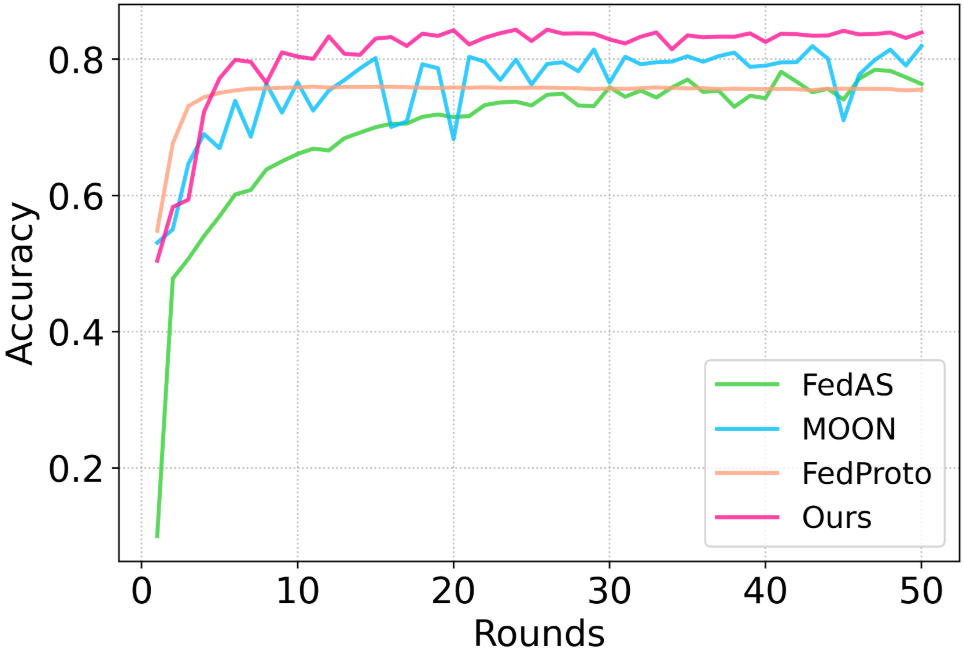}
		\captionsetup{font=small}
		\caption{EuroSAT}
		\label{fig_Communication_Rounds_EuroSAT}
	\end{subfigure} 
	\begin{subfigure}{0.245\textwidth} 
		\includegraphics[width=1 \linewidth]{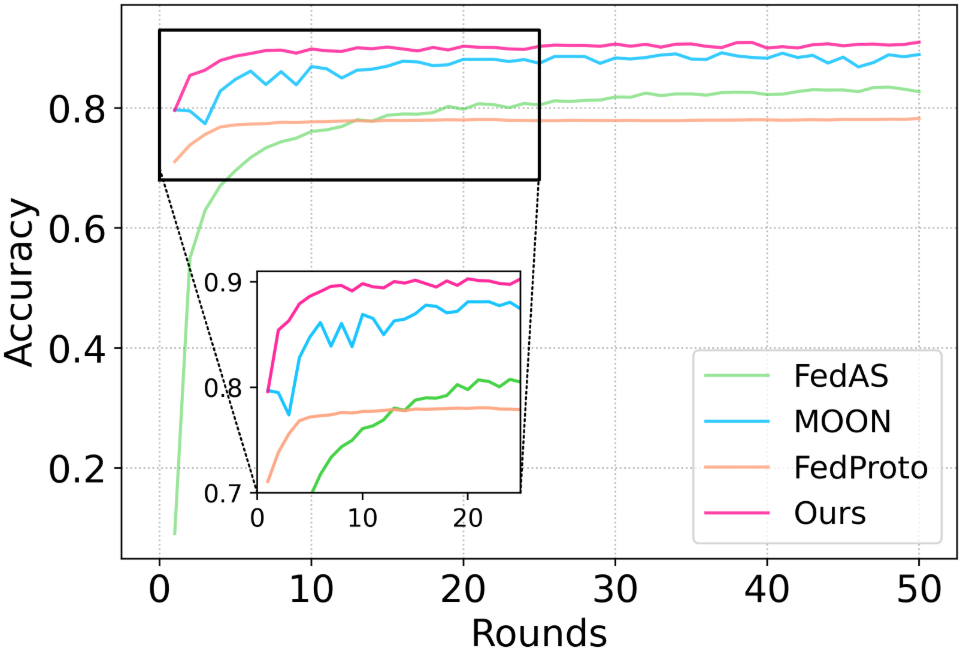}
		\captionsetup{font=small}
		\caption{Fashion-MNIST}
		\label{fig_Communication_Rounds_Fashion_MNIST}
	\end{subfigure}
	\begin{subfigure}{0.245\textwidth} 
		\includegraphics[width=1 \linewidth]{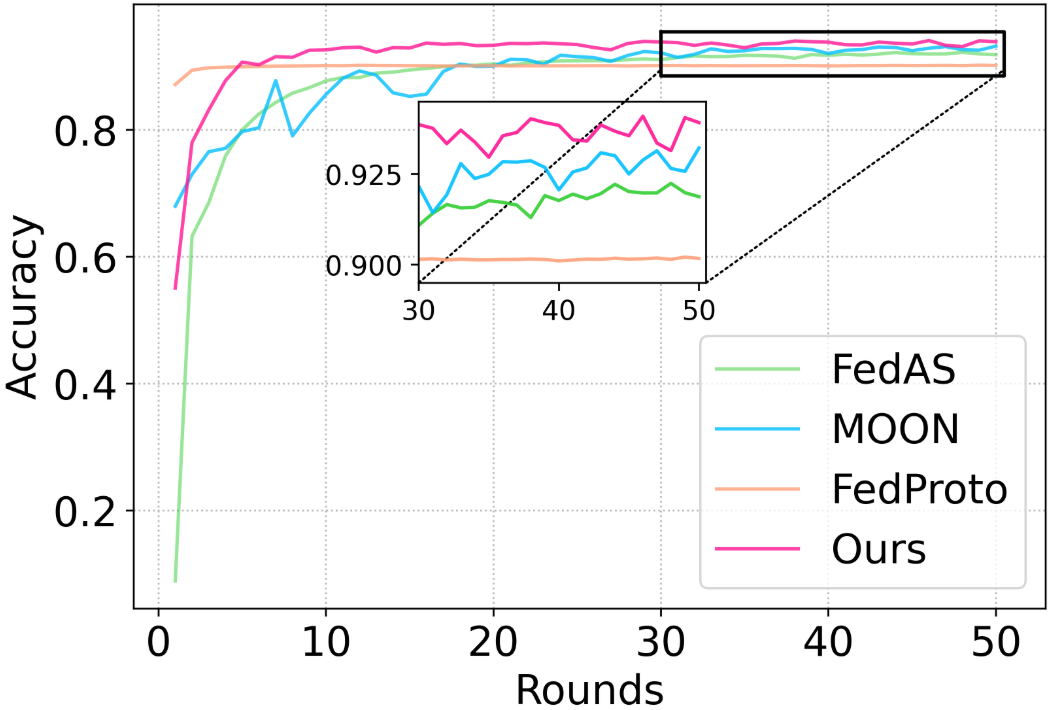}
		\captionsetup{font=small}
		\caption{M+F}
		\label{fig_Communication_Rounds_M_F}
	\end{subfigure} 
	\begin{subfigure}{0.245\textwidth} 
		\includegraphics[width=1 \linewidth]{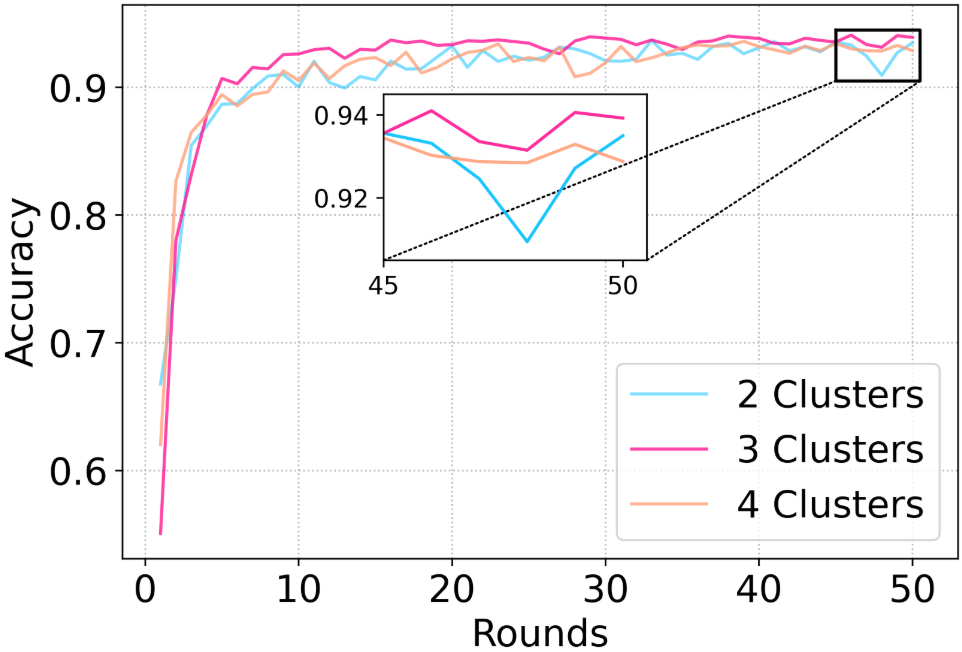}
		\captionsetup{font=small}
		\caption{Number of clusters (M+F)}
		\label{fig_Communication_Rounds_Various_Cluster_Number}
	\end{subfigure}
	\captionsetup{font=small}
	\caption {(a)-(c) Accuracy comparison among FedAS, MOON, FedProto and the proposed method via various datasets ($Dir=0.9$). (d) illustrates the accuracy achieved by the proposed method with considering different cluster number settings towards M+F dataset.}
	\label{fig_Communication_Rounds_Various_Datasets}
\end{figure*}
\begin{figure} [t!]
	\centering
	\includegraphics[scale = .42]{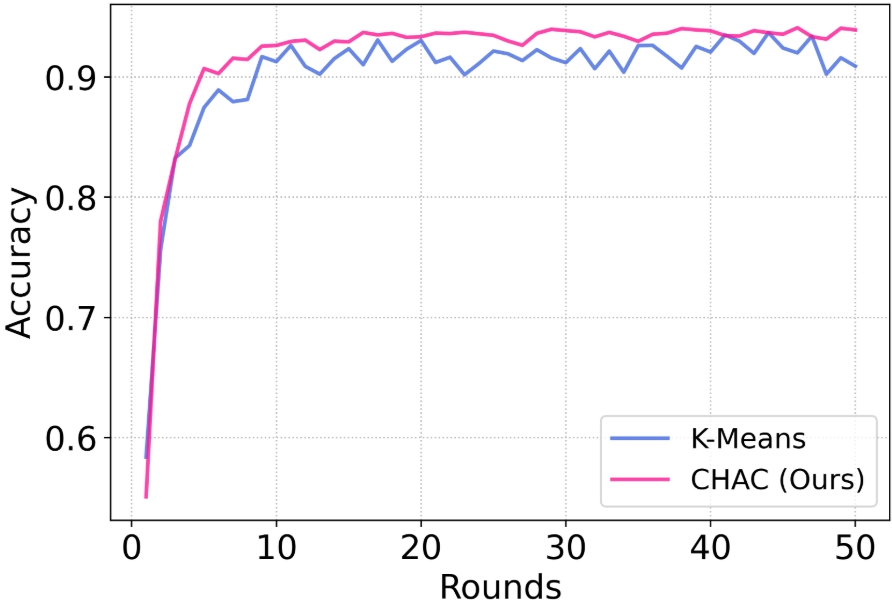}
	\caption{Accuracy versus rounds for M+F dataset. A comparison between K-Means  and CHAC based MP-FedKD is conducted.}
	\label{fig_Communication_Rounds_K_Means_Ours}
\end{figure}

To assess the \textit{\textbf{robustness}} of the proposed MP-FedKD approach, similar to \cite{L_Zou_Cyber_Attacks_Prevention_Prosumer_EV}, we also showcase the accuracy achieved in various rounds, as shown in Fig. \textcolor{red}{\ref{fig_Communication_Rounds_Various_Datasets}} and Fig. \textcolor{red}{\ref{fig_Communication_Rounds_K_Means_Ours}}.

\par
In Fig. \textcolor{red}{\ref{fig_Communication_Rounds_Various_Datasets}}, from Figs. \textcolor{red}{\ref{fig_Communication_Rounds_EuroSAT}} - \textcolor{red}{\ref{fig_Communication_Rounds_M_F}}, we compare the proposed method with FedAS, MOON and FedProto, respectively. Specifically, in the beginning, we can observe that the accuracy obtained by the proposed method is not the highest. This is reasonable because the model's training process is still in the initial exploration stage. As the number of rounds increases, this trend gradually changes. As we can see, in contrast with the considered three baselines, the proposed method can gain the most noticeable results in terms of accuracy when approaching the convergence state. For instance, consider EuroSAT as an example, in the final round, compared with the second best (i.e., MOON), the accuracy gained by the proposed method surpasses MOON by $1.98\%$. Therefore, it can be speculated that the robustness of the proposed MP-FedKD method is significant in the convergence state. In Fig. \textcolor{red}{\ref{fig_Communication_Rounds_Various_Cluster_Number}}, we manifest the accuracy obtained by the proposed method in each round under the consideration of various numbers of clusters (i.e, stopping rule for CHAC approach). It can be seen in Fig. \textcolor{red}{\ref{fig_Communication_Rounds_Various_Cluster_Number}}, the trend of accuracy is very similar under the three clustering number settings (i.e., $\{2, 3, 4\}$). Nevertheless, when the number of clusters is set as $3$, we can know from Fig. \textcolor{red}{\ref{fig_Communication_Rounds_Various_Cluster_Number}} that the achieved accuracy is the highest in the majority of cases. As a result, for the sake of simplicity, we choose $3$ as the default setting for the proposed method.

\begin{table}[t!]
	\renewcommand{\thetable}{\Roman{table}}
	\renewcommand\arraystretch{1}
	\setlength{\extrarowheight}{1 pt}
	\caption{Accuracy comparison considering various number of clients on heterogeneous datasets.}
	\begin{center}
		\begin{tabular}{c | c |>{\centering} *3{m{1.3cm}}}
			\Xhline{1pt} 
			\toprule
			\multirow{2.1}{*}{\textbf{Dataset}}& \multirow{2.1}{*}{\textbf{Method}} &\multicolumn{3}{c}{\textbf{Accuracy}} \\\cline{3-5}
			& &  \multirow{1.5}{*}{\bfseries \# 10} & \multirow{1.5}{*}{\bfseries \# 20} & \multirow{1.5}{*}{\bfseries \# 50} \\
			\midrule
			\multirow{6}{*}{M+F} & FedProx & \hfil $0.8433$  & \hfil $0.6519$ & \hfil $0.5615$ \\
			& FedAS & \hfil $0.9188$ & \hfil $0.8949$ & \hfil $0.8838$ \\
			& E-FPKD & \hfil $0.6117$ & \hfil $0.5415$ & \hfil $0.4960$ \\
			& MOON & \hfil $0.9322$ & \hfil $0.8197$ & \hfil $0.7790$ \\   
			& FedProto & \hfil $0.9041$ & \hfil $0.8342$ & \hfil $0.8133$ \\
			& \mycc Ours & \hfil \mycc $\textbf{0.9392}$ & \hfil \mycc $\textbf{0.9247}$ & \hfil \mycc $\textbf{0.9083}$ \\  
			\midrule
			\multirow{6}{*}{C+E} & FedProx & \hfil $0.3106$ & \hfil $0.3212$ & \hfil $0.2040$ \\ 
			& FedAS & \hfil $0.5654$  & \hfil $0.4601$ & \hfil $0.4855$ \\
			& E-FPKD & \hfil $0.1546$ & \hfil $0.2294$ & \hfil $0.2231$ \\
			& MOON & \hfil $0.5785$ & \hfil $0.4384$ & \hfil $0.2857$ \\  
			& FedProto  & \hfil $0.5821$ & \hfil $0.4480$ & \hfil $0.5170$ \\	
			& \mycc Ours & \hfil \mycc $\textbf{0.5897}$ & \hfil \mycc $\textbf{0.5761}$ & \hfil \mycc $\textbf{0.5705}$ \\  
			\bottomrule
			\hline
		\end{tabular}
	\end{center}
	\label{Scalability_heterogeneous_datasets}
	\vspace{-0.3cm}
\end{table}

\par
In Fig. \textcolor{red}{\ref{fig_Communication_Rounds_K_Means_Ours}}, the comparison between K-Means-based MP-FedKD and CHAC-based MP-FedKD (ours) is provided. Particularly, we observe that the K-Means-based MP-FedKD showcases notable convergence fluctuations and achieves lower accuracy in most of the rounds. When the round reaches to $50$, compared with the K-Means-based MP-FedKD, the proposed method can improve the performance by $3.02\%$. Hence, the robustness of the proposed method is remarkable compared with the K-Means-based MP-FedKD approach.
\par
Overall, from Fig. \textcolor{red}{\ref{fig_Communication_Rounds_Various_Datasets}} and Fig. \textcolor{red}{\ref{fig_Communication_Rounds_K_Means_Ours}}, it is evident that the proposed method exhibits significant robustness in addressing the considered non-IID data.

\vspace{-0.3cm}

\subsection{Ablation Study}
\par
\textbf{Various non-IID levels:} As highlighted before, non-IID data poses a challenge to the conventional FL approach. Thus, it is meaningful to study the impact of different levels of data heterogeneity on the proposed method (as shown in Table \textcolor{blue}{\ref{table_ablation_MNIST_FM_MF}}). It is worth mentioning that the smaller the $Dir$ value, the higher the level of the data heterogeneity. In particular, in Table \textcolor{blue}{\ref{table_ablation_MNIST_FM_MF}}, it can be found that the average accuracy achieved by all methods slightly decreases as the degree of heterogeneity increases. However, through the overall observation, it can be known that the proposed method achieves the highest average accuracy regardless of $Dir$ value, compared to FedProx, MOON and FedALA. Thus, we can conjecture that the proposed method demonstrates the most significant advantage in coping with heterogeneous data compared to the considered baselines.

\par
\textbf{Various parameters:} In both Tables \textcolor{blue}{\ref{table_ablation_MNIST}} and \textcolor{blue}{\ref{table_ablation_EuroSAT}}, to evaluate the efficacy of the proposed method more clearly, ablation studies in terms of different parameters, and the comparison among FedProx, MOON, FedALA, with the proposed method are provided. Concretely, we attempt to adjust the learning rate ($lr$) in the set of $\{0.1, 0.01, 0.001\}$ and the batch size ($bs$) in the set of $\{32, 64, 128\}$. The results listed in these two tables demonstrate that the proposed method can accomplish the highest average accuracy in all cases. Consider $lr=0.001, bs=32$ as a case study, the proposed method realizes the highest average accuracy with the performance gain of $0.41\%-3.04\%$ towards MNIST dataset, and with the performance gain of $4.17\%-35.88\%$ on the EuroSAT dataset, in contrast to the baselines. In addition, under this setting, we can observe that the average accuracy achieved on the MNIST dataset is the best, and on EuroSAT is the second best. Despite this situation, for convenience, we still adopt $lr=0.001, bs=32$ as the unified parameter setting.

\begin{table}[t!]
	\caption{Average Accuracy achieved by the proposed method under various Dirichlet Distribution parameters.}
	\renewcommand\arraystretch{1.2}
	\centering
	\scalebox{0.97}{
		\begin{tabular}{|c|c|cccc|}
			\Xhline{1pt}  
			\multirow{2}{*}{\textbf{Dataset}} & \multirow{2}{*}{\textbf{Method}} & \multicolumn{4}{c|}{\textbf{Average Accuracy (\# Clients=10)}}  \\ 
			\cline{3-6} 
			&  & $\textbf{Dir=0.3}$ & $\textbf{Dir=0.5}$ & $\textbf{Dir=0.7}$ & $\textbf{Dir=0.9}$ \\ \hline \hline
			\multirow{4}{*}{MNIST} & FedProx & $0.9342$ & $0.9417$ & $0.9442$ & $0.9623$  \\
			& MOON & $0.9668$ & $0.9695$ & $0.9731$ & $0.9886$  \\
			& FedALA & $0.9719$ & $0.9762$ & $0.9766$ & $0.9859$  \\ \cline{2-6}
			& Ours & $\textbf{0.9738}$ & $\textbf{0.9770}$ & $\textbf{0.9786}$ & $\textbf{0.9927}$ \\ \hline \hline
			\multirow{4}{*}{Fashion-MNIST} & FedProx & $0.6926$ & $0.7856$  & $0.7982$ & $0.8194$  \\
			& MOON & $0.8474$ & $0.8652$ & $0.8780$ & $0.8915$  \\
			& FedALA & $0.8475$ & $0.8631$ & $0.8700$ & $0.8754$  \\ \cline{2-6}
			& Ours & $\textbf{0.8710}$ & $\textbf{0.8738}$ & $\textbf{0.8834}$ & $\textbf{0.9062}$ \\ \hline \hline
			\multirow{4}{*}{M+F} & FedProx & $0.5769$ & $0.7295$  & $0.7360$  & $0.8344$ \\
			& MOON & $0.8780$ & $0.8884$ & $0.8920$ & $0.9250$ \\
			& FedALA & $0.8857$ & $0.9125$ & $0.9169$ & $0.9236$ \\ \cline{2-6}
			& Ours & $\textbf{0.8948}$ & $\textbf{0.9215}$ & $\textbf{0.9222}$ & $\textbf{0.9289}$ \\ \hline
		\end{tabular}
		\label{table_ablation_MNIST_FM_MF}
	}
\end{table}

\begin{table}[t!]
	\caption{Average Accuracy achieved by the proposed method under various learning rate and batch size (MNIST).}
	\renewcommand\arraystretch{1.2}
	\centering
	\scalebox{0.9}{
		\begin{tabular}{|p{2.6cm}|p{1.25cm}|p{1.25cm}|p{1.25cm}|p{1.3cm}|}
			\Xhline{1pt}  
			\hfil \multirow{2}{*}{\textbf{Parameter}} &  \multicolumn{4}{c|}{\textbf{Average Accuracy (\# Clients=10)}}  \\ 
			\cline{2-5} 
			& \hfil \textbf{FedProx} & \hfil \textbf{MOON} & \hfil \textbf{FedALA} & \hfil \textbf{Ours} \\ \hline 
			\hfil $lr=0.1, bs=32$ & \hfil $0.9880$ & \hfil $0.8897$ & \hfil $0.9812$ & \hfil \mycc $\textbf{0.9893}$  \\ \hline
			\hfil $lr=0.01, bs=32$ & \hfil $0.9865$ & \hfil $0.9903$ & \hfil $0.9854$ & \hfil \mycc $\textbf{0.9920}$  \\ \hline
			\hfil $lr=0.001, bs=32$ & \hfil $0.9623$ & \hfil $0.9886$ & \hfil $0.9776$ & \hfil \mycc $\textbf{0.9927}$ \\ \hline
			\hfil $lr=0.001, bs=64$ & \hfil $0.9632$ & \hfil $0.9875$ & \hfil $0.9793$ & \hfil \mycc $\textbf{0.9915}$  \\ \hline 
			\hfil $lr=0.001, bs=128$ & \hfil $0.9683$ & \hfil $0.9871$ & \hfil $0.9786$ & \hfil \mycc $\textbf{0.9881}$ \\ \hline 
		\end{tabular}
		\label{table_ablation_MNIST}
	}
\end{table}

\begin{table}[t!]
	\caption{Average Accuracy achieved by the proposed method under various learning rate and batch size (EuroSAT).}
	\renewcommand\arraystretch{1.2}
	\centering
	\scalebox{0.9}{
		\begin{tabular}{|p{2.6cm}|p{1.25cm}|p{1.25cm}|p{1.25cm}|p{1.3cm}|}
			\Xhline{1pt}  
			\hfil \multirow{2}{*}{\textbf{Parameter}} &  \multicolumn{4}{c|}{\textbf{Average Accuracy (\# Clients=10)}}  \\ 
			\cline{2-5} 
			& \hfil \textbf{FedProx} & \hfil \textbf{MOON} & \hfil \textbf{FedALA} & \hfil \textbf{Ours} \\ \hline 
			\hfil $lr=0.1, bs=32$ & \hfil $0.1040$ & \hfil $0.4329$ & \hfil $0.8191$ & \hfil \mycc $\textbf{0.8387}$ \\ \hline
			\hfil $lr=0.01, bs=32$ & \hfil $0.2649$ & \hfil $0.7750$ & \hfil $0.8185$ &  \hfil \mycc $\textbf{0.8563}$  \\ \hline
			\hfil $lr=0.001, bs=32$ & \hfil $0.4821$ & \hfil $0.7863$ & \hfil $0.7992$ &  \hfil \mycc $\textbf{0.8409}$ \\ \hline
			\hfil $lr=0.001, bs=64$ & \hfil $0.6820$ & \hfil $0.5845$ & \hfil $0.7078$ & \hfil \mycc $\textbf{0.7672}$ \\ \hline 
			\hfil $lr=0.001, bs=128$ & \hfil $0.5023$ & \hfil $0.3981$ & \hfil $0.6079$ & \hfil \mycc $\textbf{0.6702}$ \\ \hline 
		\end{tabular}
		\label{table_ablation_EuroSAT}
	}
\end{table}


\par
\textbf{Performance of PA and LEMGP loss:} In Table \textcolor{blue}{\ref{table_no_PA_no_LEMGP}}, we examine the impact of the proposed prototype alignment (PA) mechanism and the proposed LEMGP loss on the effectiveness of the proposed method in terms of average accuracy. Particularly, we train the method without using PA (w/o PA), the method without using LEMGP loss (w/o LEMGP) and the proposed method on three datsets (i.e., CIFAR10, MNIST and Fashion-MNIST), as shown in Table \textcolor{blue}{\ref{table_no_PA_no_LEMGP}}. In that Table, it can be gained that the average accuracy (AA) achieved by the proposed method is the highest among the three methods. Specifically, removing PA (i.e., w/o PA) from the proposed method results in $0.72\%$ average accuracy drop, while discard LEMGP loss (i.e., w/o LEMGP) leads to $1.58\%$ average accuracy drop on CIFAR10 dataset. Performance degradation can also be observed over the other two datasets when one of the components (PA or LEMGP loss) is detached from the proposed method. Such an interesting phenomenon can indicate that both PA and LEMGP loss are of pivotal importance for constructing the proposed method.

\begin{table}[t!]
\centering
\sffamily
\caption{Average accuracy achieved by w/o PA, w/o LEMGP and the proposed method.}
\renewcommand\arraystretch{1.2}
\begin{tabular}{||*{4}{p{1.7cm}||p{1.3cm}||p{1.2cm}||p{2.2cm}||}}
\hhline{|t:=:t:=:t:=:t:=:t|}
    \hfil \bfseries Loss  & \hfil \bfseries CIFAR-10 & \hfil \bfseries MNIST & \hfil \bfseries Fashion-MNIST \\
\hhline{|:=::=::=::=:|}
\hfil \textbf{w/o PA}  & \hfil $0.6637$   & \hfil $0.9785$   & \hfil $0.8890$         \\
\hhline{|:=::=::=::=:|}
\hfil \textbf{w/o LEMGP}   & \hfil $0.6551$  & \hfil $0.9745$  & \hfil $0.8916$    \\
\hhline{|:=::=::=::=:|}
\hfil \textbf{Ours}  & \hfil  $\textbf{0.6709}$  & \hfil $\textbf{0.9927}$  & \hfil $\textbf{0.9062}$           \\
\hhline{|b:=:b:=:b:=:b:=:b:|}                                                                              
\end{tabular}
\label{table_no_PA_no_LEMGP}
\end{table}


\section{Conclusion}
In this paper, we have proposed a method named multi-prototype-guided federated knowledge distillation (MP-FedKD) approach to deal with non-IID data issue towards AI-RAN enabled MEC system. This method involves several compositions: multi-prototype generation, self-knowledge distillation (SKD), prototype alignment, and LEMGP loss. Specifically, we have presented the conditional hierarchical agglomerative clustering (CHAC) approach for multi-prototype generation, to ameliorate the situation of useful information loss caused by using the single prototype-based strategy. For SKD, we have employed it to construct the proposed method because of its advantage in coping with the heterogeneous data and its ability to avoid preparing the teacher model beforehand. To compensate for the information loss caused by using the average manner to obtain the global prototype, we have designed a prototype alignment mechanism, in which we allow the global prototypes to learn from historical local embeddings. To enable the local embedding to align with the global prototype of the same class and separate from the global prototype of distinct class, we have designed LEMGP loss based on COREL loss. Extensive experiments manifest the practicality and effectiveness of the proposed method in handling the non-IID data issue, with improvements in accuracy and AA, along with the reduction in errors. For instance, the accuracy for CIFAR10 dataset gained by the proposed method is separately $2.01\times$, $1.50\times$, $1.82\times$, $1.48\times$ and $1.65\times$ greater than FedProx, FedAS, E-FPKD, MOON and FedProto when the number of client is set as $20$.

\vfill


\begin{thebibliography}{1}
		\bibliographystyle{IEEEtran}


        \bibitem{Z_Wu_FedICT_Federated_Multi_Task}
        Z. Wu, S. Sun, Y. Wang, M. Liu, Q. Pan, X. Jiang and B. Gao, ``FedICT: Federated Multi-Task Distillation for Multi-Access Edge Computing,'' \textit{IEEE Transactions on Parallel and Distributed Systems}, vol. 35, no. 6, pp. 1107-1121, Jun. 2024.

        \bibitem{S_D_A_Shah_Interplay_AI_RAN}
        S. D. A. Shah, Z. Nezami, M. Hafeez and S. A. R. Zaidi, ``The Interplay of AI-and-RAN: Dynamic Resource Allocation for Converged 6G Platform,'' \textit{IEEE INFOCOM 2025 - IEEE Conference on Computer Communications Workshops (INFOCOM WKSHPS)}, London, United Kingdom, pp. 1-6, May 2025.

        \bibitem{Z_Nezami_GenAI_RAN}
        Z. Nezami, S. A. R. Zaidi, M. Hafeez, J. Xu and K. Djemame, ``Towards standardization of GenAI-driven agentic architectures for radio access networks,'' \textit{Frontiers in Artificial Intelligence}, vol. 8, Jul. 2025.

        \bibitem{K_Srikandabala_Scoping}
        K. Srikandabala, K. Prabagar, S. Jayatilleke, P. Zhang, R. Ellerbrock, S. Rinas, D. D. Silva and D. Alahakoon, ``Optimization and Management of Data Center Networks: A Scoping Review on Key Themes, Challenges, and Artificial Intelligence and Machine Learning Approaches,'' \textit{IEEE Access}, vol. 13, pp. 134699-134720, Jul. 2025.


        		
		
			
		\bibitem{Y_Cheng_Pre_clustering_CFL_Data_System}
		Y. Cheng, W. Zhang, Z. Zhang, J. Kang, Q. Xu, S. Wang and D. Niyato, ``SnapCFL: A Pre-clustering-based Clustered Federated Learning Framework for Data and System Heterogeneities,'' \textit{IEEE Transactions on Mobile Computing}, vol. 24, no. 6, pp. 5214-5228, Jun. 2025.
		
		\bibitem{K_Wang_Age_of_Information_Minimization_FL}
		K. Wang, Z. Ding, D. K. C. So and Z. Ding, ``Age-of-Information Minimization in Federated Learning Based Networks With Non-IID Dataset,'' \textit{IEEE Transactions on Wireless Communications}, vol. 23, no. 8, pp. 8939-8953, Aug. 2024.

        \bibitem{Y_Xu_Overcoming_Noisy_Labels_Non_IID_Edge}
		Y. Xu, Y. Liao, L. Wang, H. Xu, Z. Jiang and W. Zhang, ``Overcoming Noisy Labels and Non-IID Data in Edge Federated Learning,'' \textit{IEEE Transactions on Mobile Computing}, vol. 23, no. 12, pp. 11406-11421, Dec. 2024.
		
		\bibitem{M_Moshawrab_Securing_FL_Approaches}
		M. Moshawrab, M. Adda, A. Bouzouane, H. Ibrahim and A. Raad, ``Securing Federated Learning: Approaches, Mechanisms and Opportunities,'' \textit{Electronics}, vol. 13, Sep. 2024.
		
		\bibitem{C_Li_FL_Soft_Clustering}
		C. Li, G. Li and P. K. Varshney, ``Federated Learning With Soft Clustering,'' \textit{IEEE Internet of Things Journal}, vol. 9, no. 10, pp. 7773-7782, May 2022.

        \bibitem{X_Li_FL_Adaptive_Weights_Non_IID}
		X. Li, Y. Gao, Y. Deng and X. Jiang, ``Federated Learning With Adaptive Aggregation Weights for Non-IID Data in Edge Networks,'' \textit{IEEE Transactions on Cognitive Communications and Networking}, vol. 11, no. 5, pp. 3425-3439, Oct. 2025.
		
		\bibitem{L_Zou_Towards_Satellite_Non_IID_Imagery}
		L. Zou, Y. M. Park, C. M. Thwal, Y. K. Tun, Z. Han and C. S. Hong, ``Towards Satellite Non-IID Imagery: A Spectral Clustering-Assisted Federated Learning Approach,'' \textit{NOMS 2025-2025 IEEE Network Operations and Management Symposium}, Honolulu, HI, pp. 1-6, May 2025.
		
		\bibitem{Z_Li_Feature_Matching_Data_Synthesis}
		Z. Li, Y. Sun, J. Shao, Y. Mao, J. H. Wang and J. Zhang, ``Feature Matching Data Synthesis for Non-IID Federated Learning,'' \textit{IEEE Transactions on Mobile Computing}, vol. 23, no. 10, pp. 9352-9367, Oct. 2024.
		
		\bibitem{K_Kenyon_Clustering_Representation}
		K. Kenyon-Dean, A. Cianflone, L. Page-Caccia, G. Rabusseau, J. C. K. Cheung and D. Precup, ``Clustering-Oriented Representation Learning with Attractive-Repulsive Loss,'' \textit{AAAI 2019 Workshop on Network Interpretability for Deep Learning}, Honolulu, Hawaii, Jan. 2019.
		
		\bibitem{J_Tang_FedRAD}
		J. Tang, X. Ding, D. Hu, B. Guo, Y. Shen, P. Ma and Y. Jiang, ``FedRAD: Heterogeneous Federated Learning via Relational Adaptive Distillation,'' \textit{Sensors}, vol. 23, Jul. 2023.
		
		
		\bibitem{L_Zou_Cyber_Attacks_Prevention_Prosumer_EV}
		L. Zou, Q. H. Vo, K. Kim, H. Q. Le, C. M. Thwal, C. Zhang and C. S. Hong, ``Cyber Attacks Prevention Towards Prosumer-Based EV Charging Stations: An Edge-Assisted Federated Prototype Knowledge Distillation Approach,'' \textit{IEEE Transactions on Network and Service Management}, vol. 22, no. 2, pp. 1972-1999, Apr. 2025.
		
		\bibitem{M_Ji_Refine_Myself_Teaching_Myself}
		M. Ji, S. Shin, S. Hwang, G. Park and I. -C. Moon, ``Refine Myself by Teaching Myself: Feature Refinement via Self-Knowledge Distillation,'' \textit{IEEE/CVF Conference on Computer Vision and Pattern Recognition (CVPR)}, Nashville, TN, pp. 10659-10668, Jun. 2021.
		
		\bibitem{H_Lee_self_knowledge_distillation}
		H. Lee, Y. Park, H. Seo and M. Kang, ``Self-knowledge distillation via dropout,'' \textit{Computer Vision and Image Understanding}, vol. 233, Aug. 2023.
		
		\bibitem{Y_Tan_FedProto}
		Y. Tan, G. Long, L. Liu, T. Zhou, Q. Lu, J. Jiang and C. Zhang, ``FedProto: Federated Prototype Learning across Heterogeneous Clients,'' \textit{Proceedings of the AAAI Conference on Artificial Intelligence}, vol. 36, No. 8, pp. 8432-8440, Jun. 2022.
		
		\bibitem{X_Mu_FedProc}
		X. Mu, Y. Shen, K. Cheng, X. Geng, J. Fu, T. Zhang, Z. Zhang, ``FedProc: Prototypical contrastive federated learning on non-IID data,'' \textit{Future Generation Computer Systems}, vol. 143, pp. 93-104, Jun. 2023.
		
		\bibitem{B_Yan_FedRFQ}
		B. Yan, H. Zhang, M. Xu, D. Yu and X. Cheng, ``FedRFQ: Prototype-Based Federated Learning With Reduced Redundancy, Minimal Failure, and Enhanced Quality,'' \textit{IEEE Transactions on Computers}, vol. 73, no. 4, pp. 1086-1098, Apr. 2024.
		
		\bibitem{Y_Tan_FL_Pre_Trained_Contrastive}
		Y. Tan, G. Long, J. Ma, L. Liu, T. Zhou and J. Jiang, ``Federated Learning from Pre-Trained Models: A Contrastive Learning Approach,'' \textit{36th Conference on Neural Information Processing Systems (NeurIPS 2022)}, New Orleans LA, pp. 19332 - 19344, Nov. 2022.

        \bibitem{G_Yan_FedVCK_Non_IID_Condensed}
        G. Yan, L. Xie, X. Gao, W. Zhang, Q. Shen, Y. Fang, and Z. Wu, ``FedVCK: Non-IID Robust and Communication-Efficient Federated Learning via Valuable Condensed Knowledge for Medical Image Analysis,'' \textit{Proceedings of the AAAI Conference on Artificial Intelligence}, Philadelphia, Pennsylvania, vol. 39, no. 20, 21904-21912, Apr. 2025.
		
		\bibitem{B_Zhang_Self_Guided_Cross_Guided_Learning}
		B. Zhang, J. Xiao and T. Qin, ``Self-Guided and Cross-Guided Learning for Few-Shot Segmentation,'' \textit{IEEE/CVF Conference on Computer Vision and Pattern Recognition (CVPR)}, Nashville, TN, pp. 8308-8317, Jun. 2021.
		
		\bibitem{J_Zhang_Dual_Expert_Distillation}
		J. Zhang, Z. Zhuang, L. Xiao, X. Wu, T. Ma and L. He, ``Dual-Expert Distillation Network for Few-Shot Segmentation,'' \textit{IEEE International Conference on Multimedia and Expo (ICME)}, Brisbane, Australia, pp. 720-725, Jul. 2023.
		
		\bibitem{L_Zou_Imbalance_Cost_Energy_Scheduling}
		L. Zou, M. S. Munir, S. S. Hassan, Y. K. Tun, L. X. Nguyen and C. S. Hong, ``Imbalance Cost-Aware Energy Scheduling for Prosumers Towards UAM Charging: A Matching and Multi-Agent DRL Approach,'' \textit{IEEE Transactions on Vehicular Technology}, vol. 73, no. 3, pp. 3404-3420, Mar. 2024.
		
		\bibitem{H_R_A_Putri_segmentation_hierarchical_clustering}
		H. R. A. Putri, A. A. R. Fernandes, A. Iriany, and S. Nurjannah, ``Credit customer segmentation with hierarchical clustering at various distances,'' \textit{Journal of Theoretical and Applied Information Technology}, vol. 101, no. 2, pp. 883–893, Jan. 2023.

        \bibitem{X_Yang_FedAS}
		X. Yang, W. Huang and M. Ye, ``FedAS: Bridging Inconsistency in Personalized Federated Learning,'' {\em IEEE/CVF Conference on Computer Vision and Pattern Recognition (CVPR)}, Seattle, WA, pp. 11986-11995, Jun. 2024.
		
		\bibitem{J_Deuschel_Multi_Prototype_Few_shot}
		J. Deuschel, D. Firmbach, C. I. Geppert, M. Eckstein, A. Hartmann, V. Bruns, P. Kuritcyn, J. Dexl, D. Hartmann, D. Perrin, T. Wittenberg and M. Benz, ``Multi-Prototype Few-shot Learning in Histopathology,'' \textit{IEEE/CVF International Conference on Computer Vision Workshops (ICCVW)}, Montreal, BC, Canada, pp. 620-628, Oct. 2021.
		
		\bibitem{A_Krizhevsky_Learning_multiple_tiny}
		A. Krizhevsky and G. Hinton, ``Learning multiple layers of features from tiny images,'' {\em Department of Computer Science, University of Toronto}, Toronto, ON, Canada, Apr. 2009.
		
		\bibitem{H_Xiao_Fashion_MNIST}
		H. Xiao, K. Rasul and R. Vollgraf, ``Fashion-MNIST: a Novel Image Dataset for Benchmarking Machine Learning Algorithms,'' {\em arXiv:1708.07747v2}, Sep. 2017.
		
		\bibitem{P_Helber_EuroSAT}
		P. Helber, B. Bischke, A. Dengel and D. Borth, ``EuroSAT: A Novel Dataset and Deep Learning Benchmark for Land Use and Land Cover Classification,'' {\em IEEE Journal of Selected Topics in Applied Earth Observations and Remote Sensing}, vol. 12, no. 7, pp. 2217-2226, Jul. 2019.
        
		\bibitem{A_H_Mohamed_Combining_Client_Selection_KD}
		A. H. Mohamed, J. B. D. D. Costa, L. A. Villas, J. C. D. Reis and A. M. D. Souza, ``Combining Client Selection Strategy with Knowledge Distillation for Federated Learning in non-IID Data,'' \textit{IEEE Symposium on Computers and Communications (ISCC)}, Paris, France, pp. 1-7, Jun. 2024.
		
		\bibitem{S_Ge_FedAMKD_Adaptive_Mutual_KD_FL}
		S. Ge, D. Liu, Y. Yang, J. He, S. Zhang and Y. Cao, ``FedAMKD: Adaptive Mutual Knowledge Distillation Federated Learning Approach for Data Quantity-Skewed Heterogeneity,'' \textit{IEEE International Conference on Systems, Man, and Cybernetics (SMC)}, Kuching, Malaysia, pp. 4710-4715, Oct. 2024.
		
	\bibitem{D_Yao_FedGKD}
		D. Yao, W. Pan, Y. Dai, Y. Wan, X. Ding, C. Yu, H. Jin, Z. Xu and L. Sun, ``FedGKD: Toward Heterogeneous Federated Learning via Global Knowledge Distillation,'' \textit{IEEE Transactions on Computers}, vol. 73, no. 1, pp. 3-17, Jan. 2024.
        

        \bibitem{Y_Sahraoui_FedRx_Distillation}
        Y. Sahraoui, C. A. Kerrache, C. T. Calafate and P. Manzoni, ``FedRx: Federated Distillation-Based Solution for Preventing Hospitals Overcrowding During Seasonal Diseases Using MEC,'' \textit{IEEE 21st Consumer Communications \& Networking Conference (CCNC)}, Las Vegas, NV, pp. 558-561, Jan. 2024.

        \bibitem{T_D_Nguyen_HFL_MEC_KD}
        T. D. Nguyen, N. A. Tong, B. P. Nguyen, Q. V. Hung Nguyen, P. L. Nguyen and T. T. Huynh, ``Hierarchical Federated Learning in MEC Networks with Knowledge Distillation,'' \textit{International Joint Conference on Neural Networks (IJCNN)}, Yokohama, Japan, pp. 1-8, Jun. 2024.       

        \bibitem{J_Chai_PFRL_AAV_MEC}
        J. Chai, Z. Wang, C. Ma, G. Gao and L. Shi, ``Personalized Federated Reinforcement Learning for Multi-AAV Assisted Edge Computing,'' \textit{IEEE Wireless Communications Letters}, vol. 14, no. 7, pp. 2074-2078, Jul. 2025.     

        \bibitem{W_Fan_Joint_Task_Offloading_MEC}
        W. Fan, J. Han, Y. Su, X. Liu, F. Wu, B. Tang and Y. Liu, ``Joint Task Offloading and Service Caching for Multi-Access Edge Computing in WiFi-Cellular Heterogeneous Networks,'' \textit{IEEE Transactions on Wireless Communications}, vol. 21, no. 11, pp. 9653-9667, Nov. 2022.

         \bibitem{G_Sun_FeDistSlice_Federated_Policy}
        G. Sun, D. Ayepah-Mensah, H. Chen, G. O. Boateng and G. Liu, ``FeDistSlice: Federated Policy Distillation for Collaborative Intelligence in Multi-Tenant RAN Slicing,'' \textit{IEEE Transactions on Services Computing}, vol. 18, no. 1, pp. 184-197, Jan.-Feb. 2025.

        \bibitem{T_Kwantwi_PFL_MEC_RAN}
        T. Kwantwi, G. Sun, N. A. E. Kuadey, G. T. Maale and G. Liu, ``Personalized Federated Learning for Intelligent Slice-Based Task Offloading and Slice Resource Allocation in Sliced B5G MEC-Enabled Network,'' \textit{IEEE Internet of Things Journal}, vol. 12, no. 23, pp. 49992-50011, Dec. 2025.

        \bibitem{Z_Wang_Communication_Efficient_PFL_DT}
        Z. Wang, X. Ma, H. Zhang and D. Yuan, ``Communication-Efficient Personalized Federated Learning for Digital Twin in Heterogeneous Industrial IoT,'' \textit{IEEE International Conference on Communications Workshops (ICC Workshops)}, Rome, Italy, pp. 237-241, May 2023.

        \bibitem{G_Liao_PFL_SKD_VEC}
        G. Liao, Y. Yang and Z. Feng, ``Personalized federated learning through self-knowledge distillation in vehicular edge computing,'' \textit{Computer Networks}, vol. 269, Sep. 2025.
		

       \bibitem{T_Gao_FedPC}
       T. Gao, K. Liu, Y. Yang, X. Liu, P. Zhang and G. Wang, ``FedPC: An Efficient Prototype-Based Clustered Federated Learning on Medical Imaging,'' \textit{IEEE Journal of Biomedical and Health Informatics}, vol. 29, no. 10, pp. 7396-7408, Oct. 2025.
       
      \bibitem{L_Chai_Prototype_based_fine_tuning_mitigating}
       L. Chai, J. Xie and N. Zhou, ``Prototype-based fine-tuning for mitigating data heterogeneity in federated learning,'' \textit{Future Generation Computer Systems}, vol. 170, Sep. 2025.

       \bibitem{Y_Zhou_FedSA}
       Y. Zhou, X. Qu, C. You, J. Zhou, J. Tang, X. Zheng, C. Cai and Y. Wu, ``FedSA: A Unified Representation Learning via Semantic Anchors for Prototype-based Federated Learning,'' \textit{Proceedings of the AAAI Conference on Artificial Intelligence}, Philadelphia, Pennsylvania, vol. 39, no. 21, pp. 23009-23017, Apr. 2025.

       \bibitem{T_K_Tran_FedNTProto}
       T. -K. Tran, H. -P. Tran, T. -L. Le and T. -H. Tran, ``FedNTProto: A Prototype-Based Approach for Personalized Federated Learning,'' \textit{International Conference on Multimedia Analysis and Pattern Recognition (MAPR)}, Da Nang, Vietnam, Aug. 2024.

       \bibitem{H_Li_FedCPG}
       H. Li, X. Wang, P. Cao, Y. Li, B. Yi and M. Huang, ``FedCPG: A class prototype guided personalized lightweight federated learning framework for cross-factory fault detection,'' \textit{Computers in Industry}, vol. 164, Jan. 2025.
                
	    \bibitem{L_Wang_Taming_Cross}
		L. Wang, J. Bian, L. Zhang, C. Chen and J. Xu, ``Taming Cross-Domain Representation Variance in Federated Prototype Learning with Heterogeneous Data Domains,'' \textit{38th Conference on Neural Information Processing Systems (NeurIPS 2024)}, Vancouver, Canada, pp. 88348 - 88372, Dec. 2024.
							
		
	\bibitem{X_Xu_Multiple_Adaptive_Prototypes_PFL}
		X. Xu, C. Zhao, L. Zhu and J. Zhang, ``Multiple Adaptive Prototypes Learning in Personalized Federated Learning,''  \textit{China Automation Congress (CAC)}, Chongqing, China, pp. 2347-2352, Nov. 2023.

        \bibitem{Y_Bi_Multi_Prototype_Embedding_Refinement}
        Y. Bi, E. Che, Y. Chen, Y. He and J. Qu, ``Multi Prototype Embedding Refinement for Semi-Supervised Medical Image Segmentation,''  \textit{arXiv:2503.14343}, Mar. 2025.

        \bibitem{M_Le_FedMP_Multi_Prototype}
        M. Le, Z. Li, C. Liu, G. Gou, G. Xiong and W. Xia, ``FedMP: Robust and Communication-Efficient Federated Multi-Prototype Intrusion Detection Framework in IoT,''  \textit{IEEE 29th International Conference on Parallel and Distributed Systems (ICPADS)}, Ocean Flower Island, China, pp. 884-891, Dec. 2023.

        \bibitem{K_Zhang_Federated_Learning_Heterogeneous_GDBD}
        K. Zhang, J. Wang, W. Wang, T. Zeng, P. Li, X. Wang, and T. Zhang, ``Federated learning with heterogeneous data and models based on global decision boundary distillation,''  \textit{Journal of King Saud University Computer and Information Sciences}, vol. 37, Jun. 2025.


        \bibitem{S_F_Peng_Multi_Granularity_Aggregation_Network}
        S. -F. Peng, G. -S. Xie, F. Zhao, X. Shu and Q. Liu, ``Multi-Granularity Aggregation Network for Remote Sensing Few-Shot Segmentation,''  \textit{IEEE Transactions on Geoscience and Remote Sensing}, vol. 63, pp. 1-14, 2025.

        \bibitem{R_Fan_NAPG}
        R. Fan, J. Xie, J. Liu, J. Zhang, Y. Zhang, H. Hou and J. Yang, ``NAPG: Neighborhood-Assisted Multiprototype Group Model for Cross-Domain Semantic Segmentation of Remote Sensing Images,''  \textit{IEEE Transactions on Geoscience and Remote Sensing}, vol. 63, pp. 1-19, Jul. 2025.
        		
	    \bibitem{S_Sarfraz_FINCH}
		S. Sarfraz, V. Sharma and R. Stiefelhagen, ``Efficient Parameter-Free Clustering Using First Neighbor Relations,'' \textit{IEEE/CVF Conference on Computer Vision and Pattern Recognition (CVPR)}, Long Beach, CA, pp. 8926-8935, Jun. 2019.

        \bibitem{A_Younis_Communication_Efficient_NG_RAN}
        A. Younis, C. Sun and D. Pompili, ``Communication-Efficient Disaggregated and Distributed Federated Learning in NG-RANs,'' \textit{IEEE Transactions on Network and Service Management}, vol. 22, no. 6, pp. 5155-5166, Dec. 2025.
        
		
		\bibitem{H_B_McMahan_Communication_Decentralized}
		H. B. McMahan, E. Moore, D. Ramage, S. Hampson and B. Agueray Arcas, ``Communication-Efficient Learning of Deep Networks from Decentralized Data,'' \textit{Proceedings of the 20th International Conference on Artificial Intelligence and Statstics}, Fort Lauderdale, FL, pp. 1273-1282, Apr. 2017.
		
		\bibitem{L_Zou_EFCKD}
		L. Zou, H. Q. Le, A. D. Raha, D. U. Kim and C. S. Hong, ``EFCKD: Edge-Assisted Federated Contrastive Knowledge Distillation Approach for Energy Management: Energy Theft Perspective,'' \textit{24st Asia-Pacific Network Operations and Management Symposium (APNOMS)}, Sejong, Korea, Republic of, pp. 30-35, Sep. 2023.
		
		\bibitem{Z_Zhu_ISFL_FL_Non_IID_Importance}
		Z. Zhu, Y. Shi, P. Fan, C. Peng and K. B. Letaief, ``ISFL: Federated Learning for Non-i.i.d. Data With Local Importance Sampling,'' \textit{IEEE Internet of Things Journal}, vol. 11, no. 16, pp. 27448-27462, Aug. 2024.

        

        \bibitem{H_Jin_pFedSD}
		H. Jin, D. Bai, D. Yao, Y. Dai, L. Gu, C. Yu and L. Sun, ``Personalized Edge Intelligence via Federated Self-Knowledge Distillation,'' \textit{IEEE Transactions on Parallel and Distributed Systems}, vol. 34, no. 2, pp. 567-580, Feb. 2023.
		
		\bibitem{B_Peng_Correlation_Congruence_ICCV}
		B. Peng, X. Jin, J. Liu, D. Li, Y. Wu, Y. Liu, S. Zhou and Z. Zhang, ``Correlation Congruence for Knowledge Distillation,'' \textit{Proceedings of the IEEE/CVF International Conference on Computer Vision (ICCV)}, pp. 5007-5016, Seoul, South Korea, Oct. 2019.
		
		\bibitem{G_Hinton_Distilling_Knowledge_NN}
		G. Hinton, O. Vinyals and J. Dean, ``Distilling the Knowledge in a Neural Network,'' \textit{Deep Learning and Representation Learning Workshop: NIPS}, Montreal, Quebec, Canada, Dec. 2014.
		
		\bibitem{S_Hirano_Comparison_clustering}
		S. Hirano, X. Sun and S. Tsumoto, ``Comparison of clustering methods for clinical databases,'' {\em Information Sciences}, vol. 159, pp. 155-165, Feb. 2004. 
		
		\bibitem{C_Hervada_Sala_program}
		C. Hervada-Sala and E. Jarauta-Bragulat, ``A program to perform Ward’s clustering method on several regionalized variables,'' {\em Computers \& Geosciences}, vol. 30, pp. 881-886, Oct. 2004.
		
		\bibitem{R_R_Yager_Intelligent_control}
		R. R. Yager, ``Intelligent control of the hierarchical agglomerative clustering process,'' {\em IEEE Transactions on Systems, Man, and Cybernetics, Part B (Cybernetics)}, vol. 30, no. 6, pp. 835-845, Dec. 2000.
		
		
        \bibitem{P_Khosla_Supervised_Contrastive}
        P. Khosla,  P. Teterwak, C. Wang, A. Sarna, Y. Tian, P. Isola, A. Maschinot, C. Liu and D. Krishnan, ``Supervised Contrastive Learning,'' {\em Advances in Neural Information Processing Systems 33 (NeurIPS 2020)}, virtual, pp. 18661 - 18673, Dec. 2020.
		
		\bibitem{X_Li_FedGTA}
		X. Li, Z. Wu, W. Zhang, Y. Zhu, R. Li and G. Wang, ``FedGTA:
		Topology-aware Averaging for Federated Graph Learning,'' {\em Proceedings of the VLDB Endowment}, vol. 17, pp. 41-50, Sep. 2023.
		
		\bibitem{S_Sieranoja_Fast_agglomerative_clustering}
		S. Sieranoja and P. Fränti, ``Fast agglomerative clustering using approximate traveling salesman solutions,'' {\em Journal of Big Data}, vol. 12, no. 21, Jan. 2025.
		
		\bibitem{K_He_CNN_Constrained_Time}
		K. He and J. Sun, ``Convolutional Neural Networks at Constrained Time Cost,'' {\em Proceedings of the IEEE Conference on Computer Vision and Pattern Recognition}, Boston, MA, pp. 5353-5360, Jun. 2015.
		

        \bibitem{C_Pan_Fair_Graph_FL}
		C. Pan, J. Xu, Y. Yu, Z. Yang, Q. Wu, C. Wang, L. Chen and Y. Yang,``Towards Fair Graph Federated Learning via Incentive Mechanisms,'' {\em arXiv:2312.13306}, Dec. 2023.
        
				
		\bibitem{Demystifying_Impact_Key_FL}
		M. Kundroo and T. Kim, ``Demystifying Impact of Key Hyper-Parameters in Federated Learning: A Case Study on CIFAR-10 and FashionMNIST,'' {\em IEEE Access}, vol. 12, pp. 120570-120583, Aug. 2024.
		
		\bibitem{Z_Liu_Privacy_Split_Learning_Ensembles}
		Z. Liu, J. Guo, M. Yang, W. Yang, J. Fan and K. Lam, ``Privacy-Enhanced Knowledge Transfer with Collaborative Split Learning over Teacher Ensembles,'' {\em SecTL '23: Proceedings of the Secure and Trustworthy Deep Learning Systems Workshop}, Melbourne VIC Australia, Jul. 2023.
		
		\bibitem{T_Liang_Compressing_Multiobject}
		T. Liang, M. Wang, J. Chen, D. Chen, Z. Luo and and V. C. M. Leung, ``Compressing the Multiobject Tracking Model via Knowledge Distillation,'' {\em IEEE Transactions on Computational Social Systems}, vol. 11, no. 2, pp. 2713-2723, Apr. 2024.
		
		\bibitem{T_Li_FedProx}
		T. Li, A. K. Sahu, M. Zaheer, M. Sanjabi, A. Talwalkar, V. Smith, ``Federated Optimization in Heterogeneous Networks,'' {\em Proceedings of Machine Learning and Systems 2}, Austin, TX, pp. 429-450, Mar. 2020.
		
		\bibitem{Q_Li_MOON}
		Q. Li, B. He, and D. Song, ``Model-Contrastive Federated Learning,'' {\em Proceedings of the IEEE/CVF Conference on Computer Vision and Pattern Recognition}, Nashville, TN, pp. 10713-10722, Jun. 2021.
		
		\bibitem{J_Zhang_FedALA}
		J. Zhang, Y. Hua, H. Wang, T. Song, Z. Xue, R. Ma and H. Guan, ``FedALA: Adaptive Local Aggregation for Personalized Federated Learning,'' {\em Proceedings of the AAAI Conference on Artificial Intelligence}, Washington DC, USA, vol. 37, no. 9, pp. 11237-11244, Jun. 2023.
		
		\bibitem{J_Park_Development_WEEE}
		J. Park, K. V. Park, S. Yoo, S. O. Choi and S. W. Han, ``Development of the WEEE grouping system in South Korea using the
		hierarchical and non-hierarchical clustering algorithms,'' {\em Resources, Conservation \& Recycling}, vol. 161, Oct. 2020.
		
		\bibitem{M_Abdel_Basset_Lightweight_Convolutional}
		M. Abdel-Basset, H. Hawash, K. M. Sallam, I. Elgendi, K. Munasinghe and A. Jamalipour, ``Efficient and Lightweight Convolutional Networks for IoT Malware Detection: A Federated Learning Approach,'' {\em IEEE Internet of Things Journal}, vol. 10, no. 8, pp. 7164-7173, Apr. 2023.
		
	\end{thebibliography}
\end{document}